\documentclass[lettersize,journal]{IEEEtran}
\usepackage{amsmath,amsfonts,amssymb}
\usepackage[ruled, linesnumbered]{algorithm2e}
\usepackage{setspace}
\usepackage{array}
\usepackage[caption=false,font=normalsize,labelfont=sf,textfont=sf]{subfig}
\usepackage{textcomp}
\usepackage{stfloats}
\usepackage{url}
\usepackage{verbatim}
\usepackage{graphicx}
\usepackage{cite}

\usepackage{tikz}
\usepackage{comment}
\usepackage{color}
\usepackage{amsfonts}
\usepackage{bm}
\usepackage{booktabs}
\usepackage{tabu}
\usepackage{multirow}
\usepackage{utfsym}
\usepackage{bbding}
\def\etal{\emph{et al.}}
\def\ie{\emph{i.e.},}

\begin{document}

\title{Reliability-Hierarchical Memory Network for Scribble-Supervised Video Object Segmentation}

\author{Zikun~Zhou,
        Kaige~Mao,
        Wenjie~Pei,
        Hongpeng~Wang,\\
        Yaowei~Wang,~\IEEEmembership{Member, IEEE,}
        and Zhenyu~He,~\IEEEmembership{Senior Member, IEEE}
\thanks{
    Zikun Zhou and Yaowei Wang are with Peng Cheng Laboratory, Shenzhen, China (e-mail: zhouzikunhit@gmail.com; wangyw@pcl.ac.cn). Zikun Zhou is also with the School of Computer Science and Technology, Harbin Institute of Technology, Shenzhen, China.

    Kaige Mao, Wenjie Pei, Hongpeng Wang, and Zhenyu He are with the School of Computer Science and Technology, Harbin Institute of Technology, Shenzhen, China (e-mail: maokaige.hit@gmail.com; wenjiecoder@outlook.com; \{wanghp; zhenyuhe\}@hit.edu.cn). Hongpeng Wang and Zhenyu He are also with Peng Cheng Laboratory, Shenzhen, China.
    
    Zikun Zhou and Kaige Mao contribute equally to this work.
    
    Yaowei Wang and Zhenyu He are the corresponding authors.
    }
}

\markboth{Journal of \LaTeX\ Class Files,~Vol.~14, No.~8, August~2021}%
{Shell \MakeLowercase{\textit{et al.}}: A Sample Article Using IEEEtran.cls for IEEE Journals}


\maketitle

\begin{abstract}
This paper aims to solve the video object segmentation (VOS) task in a scribble-supervised manner, in which VOS models are not only trained by the sparse scribble annotations but also initialized with the sparse target scribbles for inference. Thus, the annotation burdens for both training and initialization can be substantially lightened. The difficulties of scribble-supervised VOS lie in two aspects. On the one hand, it requires the powerful ability to learn from the sparse scribble annotations during training. On the other hand, it demands strong reasoning capability during inference given only a sparse initial target scribble. In this work, we propose a Reliability-Hierarchical Memory Network (RHMNet) to predict the target mask in a step-wise expanding strategy w.r.t. the memory reliability level. To be specific, RHMNet first only uses the memory in the high-reliability level to locate the region with high reliability belonging to the target, which is highly similar to the initial target scribble. Then it expands the located high-reliability region to the entire target conditioned on the region itself and the memories in all reliability levels. Besides, we propose a scribble-supervised learning mechanism to facilitate the learning of our model to predict dense results. It mines the pixel-level relation within the single frame and the frame-level relation within the sequence to take full advantage of the scribble annotations in sequence training samples. The favorable performance on two popular benchmarks demonstrates that our method is promising.
\end{abstract}

\begin{IEEEkeywords}
Video object segmentation, weakly supervised, scribble, memory network, hierarchical reliability.
\end{IEEEkeywords}

\section{Introduction}
\vspace{-1mm}
\IEEEPARstart{V}{ideo} Object Segmentation (VOS) is the task aiming to predict the pixel-level mask for every target object in every frame of a video. It has a wide range of applications, including robotic manipulation~\cite{Robot-Manipulation,Robot-Grasps}, video surveillance~\cite{Surveillance,Camera, URVOS-TIP}, and video editing~\cite{VOS-RMP}, and has attracted more and more attention in the computer vision community. In this paper, we focus the VOS task in the one-shot setting, in which the annotation of the target object in the first frame is provided during inference\footnote{The VOS task in the one-shot setting is also called semi-supervised VOS in literature.}. We mean the VOS task in the one-shot setting when referring to VOS for presentation clarity.

Recently, the VOS field has witnessed astonishing progress with the development of powerful deep learning-based methods~\cite{LITL-VOS,STM-VOS,PClip-VOS,XMem-VOS,Unicorn}. However, most of the existing methods require a mask annotation representing the target object to initialize the VOS model. It is less user-friendly to require users to provide an accurate mask annotation, which takes dozens of seconds to annotate, for every target object in practical applications. To alleviate this issue, LWL~\cite{LWL-VOS} introduces a bounding box encoder to convert the initial bounding box annotation to a mask, enabling the model to use the bounding box for initialization. Converting bounding boxes to masks to initialize the VOS model originally designed for initializing with manually annotated accurate masks is not a neat way to solve this problem. We believe the potential of using weak annotations for initialization has not been well explored. 

\begin{figure}[t]
\centering
    \includegraphics[width=0.98\columnwidth]{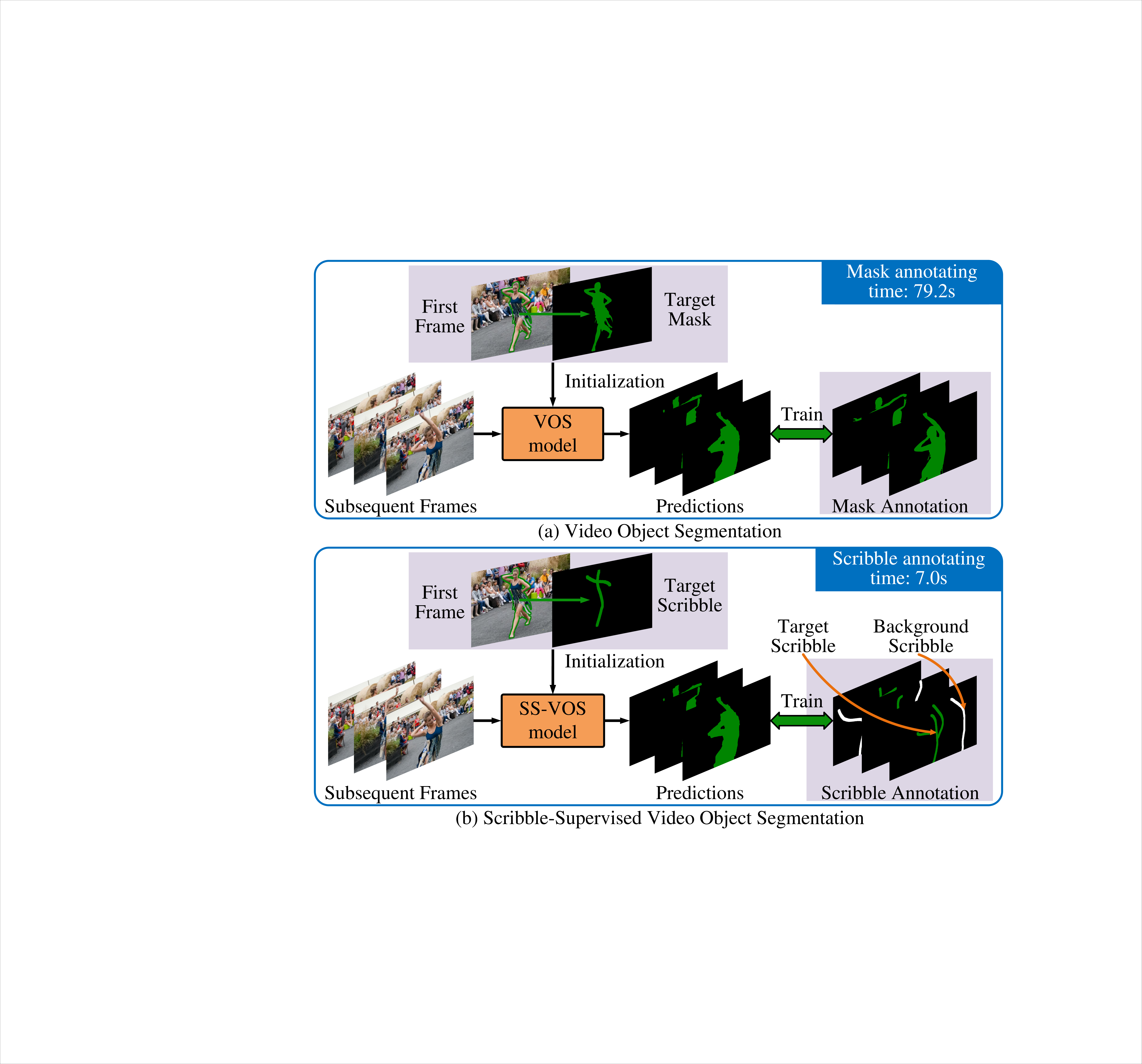}\vspace{-1mm}
    \caption{Unlike the conventional VOS task in which mask annotations are available, the SS-VOS task only provides scribble annotations for model training and initialization. Compared with labeling mask annotation, which takes 79.2 seconds per object with annotation tools as reported in COCO, drawing scribble annotation with a pencil takes only 7.0 seconds per object and is more user-friendly and time-saving.}
\label{Fig:Introduction}
\end{figure}
On the other hand, most methods~\cite{STM-VOS,KMN-VOS,RANet,AOTrans} rely on a large amount of pixel-wise annotations, \ie~masks, to learn to track and segment the target objects in a video. Annotating pixel-wise masks in a large number of videos is an extremely time-consuming and laborious task. 
One of the solutions to lighten the annotation burden for training is to resort to weakly supervised learning, such as training the VOS model with the bounding box or scribble annotations. The weakly supervised learning technique has been widely explored in many vision tasks, e.g., object detection~\cite{WSDDet,ResWSOD} and semantic segmentation~\cite{Scribblesup,CPWSSS}, demonstrating the great potential for utilizing the weakly supervised annotation.

An intuitive idea for lightening annotation burdens in initialization and training is to solve this task in a weakly supervised manner. That is, the VOS model is not only trained but also initialized with weak annotations. Boxes and scribbles are common user-friendly weak annotations for interacting, taking only a few seconds for drawing. Considering that scribbles coarsely depict the object skeleton and represent the winding object with less ambiguity than boxes, we opt for scribbles in this work. Using scribbles for training and initializing, the conventional VOS problem becomes the Scribble-Supervised Video Object Segmentation (SS-VOS) problem, as shown in Figure~\ref{Fig:Introduction}. 
Huang \etal~\cite{SSVOS} recently dove into the SS-VOS task, and they introduced a scribble attention module and a scribble-supervised loss into a propagation-based VOS model~\cite{GAM-VOS} to enable it to use scribble annotations for initializing and training, respectively. However, the algorithm~\cite{SSVOS} still uses the propagation method originally designed for conventional VOS to conduct SS-VOS, overlooking the sparsity of the scribble. Herein the sparsity means only part target information is known and increases the difficulty to segment the target accurately. As a result, this algorithm~\cite{SSVOS}, which directly propagates the mask predicted based on the sparse scribble frame by frame, suffers from severe error accumulation issues and even collapses.

In this work, we propose a Reliability-Hierarchical Memory Network (RHMNet) for SS-VOS to alleviate the error accumulation issue. RHMNet adopts a step-wise expanding strategy w.r.t the memory reliability level to alleviate the effect of historical segmentation errors. Specifically, RHMNet maintains a hierarchical memory bank that separately stores the historical information in three reliability levels: 1) the initial scribble, 2) the region highly similar to the initial scribble (named reliable region), and 3) the entire target mask. As the cover range dilates from the scribble to the entire target, the reliability of the corresponding historical information decreases. 
Thus, RHMNet only uses historical information in the corresponding or higher reliability levels as the reference for each expanding step. In particular, RHMNet processes a new frame in two steps: 1) capturing the reliable region only according to the initial scribble and the reliable regions in previous frames, and 2) using the captured reliable region and all memory information as the reference to segment the entire target.

To learn to predict the reliable region and the entire target with sparse scribble annotations, we propose a scribble-supervised learning mechanism. On the one hand, we directly extend the target scribbles to dense supervision signals for learning to segment the reliable region. On the other hand, we excavate the pixel-level relation within the single image and the frame-level relation within the sequence to supervise the learning for segmenting the entire target. To be specific, we resort to the smoothness constraint to exploit the pixel-level relation, which requires similar adjacent pixels to have similar predictions. We propose a bidirectional prediction consistency constraint to exploit the frame-level relation, which is the appearance variation across frames.

To assess the performance of our approach, we synthesize scribble annotations that mimic human annotations to train our model, and evaluate our method on DAVIS~\cite{DAVIS} and Youtube-VOS~\cite{Youtube-vos} using manually drawn scribbles for initialization. The favorable performance demonstrates that our method is promising. Our contributions are summarized as follows:
\begin{itemize}
    \item We propose a reliability-hierarchical memory network for SS-VOS; it performs step-wise expanding w.r.t. the memory reliability level to enhance the robustness against the previous prediction errors.
    \item We propose a scribble-supervised learning mechanism to effectively train the model to predict accurate dense masks with only sparse scribble annotations.
    \item We evaluate the proposed method on the popular benchmarks, and the favorable performance demonstrates the potential of our method for addressing the SS-VOS task.
    
\end{itemize}
\vspace{-3mm}

\section{Related work}
We mainly discuss the closely related studies in video object segmentation and scribble-supervised segmentation.

\subsection{Video object segmentation.} 
Most recent VOS methods could be broadly divided into two categories: propagation-based methods and matching-based methods. The propagation-based methods~\cite{SAT-VOS,MaskRCNN,GAM-VOS,RAMP-VOS,VOS-RMP, LFSI-VOS,TA-VOS} propagate the target masks in the previous frame to the current frame and then use a  network to refine the propagated mask. The matching-based methods~\cite{VM-VOS,DTM-VOS,PLM-VOS,CFBI-VOS, ASRF-TIP, DMM-VOS, TacklingBD} perform pixel-by-pixel matching between a certain previous frame and the current frame in an embedding space to predict the target mask. To improve the adaptive ability to target appearance variations, several algorithms~\cite{EGMN-VOS,STM-VOS, MaskVOS-TIP, AOLM-TIP, HMMNet-VOS, QADM-VOS, SWEM} equip the matching-based method with a memory storing the predicted target masks to exploit the long-term historical target information.

Although obtaining astonishing performance, the above methods require a mask annotation of the target for initialization, which is less user-friendly. To overcome the limitation, a few methods~\cite{LWL-VOS,QMRA-VOS, SiamRCNN,SiamMask} explore the VOS models that can be initialized with weak annotations, such as bounding boxes. SiamRCNN~\cite{SiamRCNN}, E3SN~\cite{E3SN}, and QMA~\cite{QMRA-VOS} predict the target mask in two steps on every frame: first predicting the target bounding box, and then segmenting the mask within the predicted bounding box. LWL~\cite{LWL-VOS} solves this problem by learning a bounding box encoder to convert the initial bounding box to a mask, which is further used to initialize the VOS model. Unlike these methods using the bounding box for initialization, we explore using the scribble for initialization, which represents the winding object better. Note that our approach is also different from the interactive VOS methods~\cite{MIVOS,UGVOS}. These methods~\cite{MIVOS,UGVOS} require users to repeatedly provide scribble annotations for multi-round mask refinement, while ours just need users to annotate the initial scribble for one single time.

Different from the above methods requiring mask annotations for offline training, Huang \etal~\cite{SSVOS} recently proposed an algorithm for the SS-VOS task, in which the VOS model is trained and initialized with scribble annotations. Our algorithm is also designed for the SS-VOS task but is different from~\cite{SSVOS} in two aspects: 1) Huang \etal~\cite{SSVOS} utilize a scribble attention module and a scribble-supervised loss to modify A-GAME~\cite{GAM-VOS}, a propagation-based conventional VOS model, to an SS-VOS model. This work~\cite{SSVOS} overlooks the sparsity of the scribble annotation and consequently suffers from severe error accumulation issues. By contrast, our method, which is originally designed for SS-VOS, mitigates the error accumulation issue via the step-wise expanding strategy w.r.t the memory reliability level. 2) Huang \etal~\cite{SSVOS} only consider the low-level information in the single image, including pixel color and 
spatial position, for imposing dense supervision, while our approach also excavates the frame-level relation in videos to impose dense supervision.

\begin{figure*}[t]
\centering
    \includegraphics[width=1.0\textwidth]{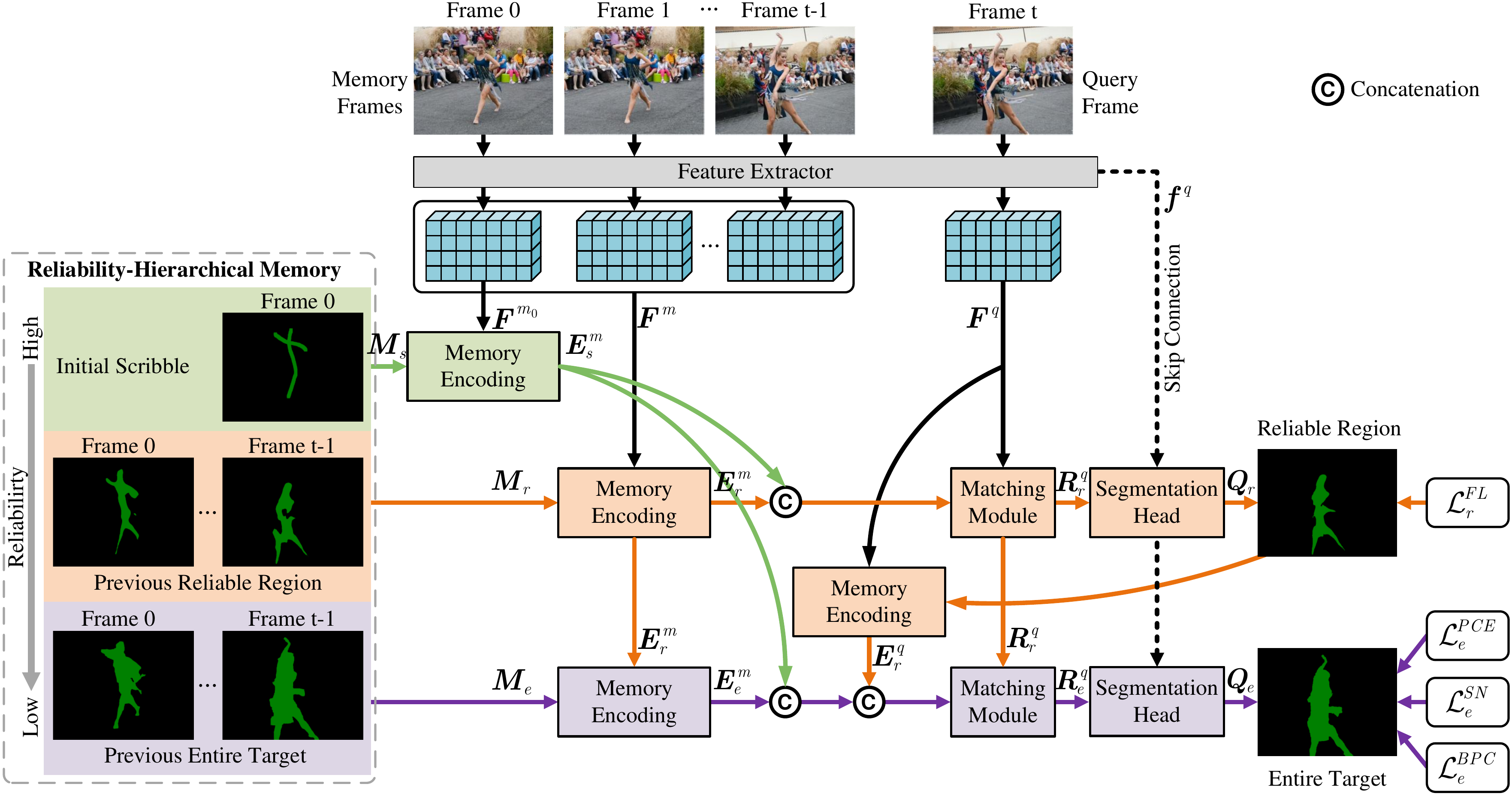}
    \vspace{-4mm}
    \caption{Overall framework of our Reliability-Hierarchical Memory Network, which consists of the reliability-hierarchical memory bank, the feature extractor, the memory encoding module, the matching module, and the segmentation head. For processing a new frame, the proposed method first captures the reliable region, which is the region highly similar to the initial target scribble region, and then accordingly segments the entire target. In each expanding step, only the historical information in the corresponding or higher reliability level is used as the reference for memory matching.}
\label{Fig:Framework}
\end{figure*}
\subsection{Scribble-supervised semantic segmentation.}
Scribble-supervised learning has been explored in some other segmentation tasks, such as semantic segmentation~\cite{Scribblesup,GatedCRF-WSSS,NormalizedCut-WSSS,KernelCut-WSSS,BPG-WSSS,DFR-WSSS}, salient object detection~\cite{SCWSSOD,WSSA}, and video salient object detection~\cite{WSVSOD}. ScribbleSup~\cite{Scribblesup} is the first method to tackle the scribble-supervised semantic segmentation, which introduces a graph model to propagate the scribble information and learn network parameters. To improve the segmentation performance, different regularized losses~\cite{GatedCRF-WSSS,NormalizedCut-WSSS,KernelCut-WSSS,DFR-WSSS} are proposed for scribble-supervised learning. Besides, the boundary detection modules~\cite{BPG-WSSS,WSSA} are introduced to extract structural information from images for predicting more precise segmentation masks. In the scribble-supervised video salient object detection task, WSVSOD~\cite{WSVSOD} designs a foreground-background similarity loss to utilize the labeling similarity across frames. Although scribble-supervised learning has been broadly studied in semantic segmentation and salient object detection, it has received less attention in the video object segmentation research community.

\section{Method}
The crux of SS-VOS is only sparse annotations are available for training and initialization while aiming to predict dense masks for the changing target. To surmount this crux, we propose a Reliability-Hierarchical Memory Network (RHMNet) to predict the target mask via step-wise expanding w.r.t. the memory reliability level. This design enables our model to utilize rich historical information and meanwhile alleviates the sensitivity to historical prediction errors. We also introduce a smooth constraint loss and a bidirectional prediction consistency loss to impose dense supervision.

\subsection{Reliability-hierarchical step-wise expanding framework}
\label{Sec: Framework}
As shown in Figure~\ref{Fig:Framework}, our RHMNet maintains a reliability-hierarchical memory of historical target states, including three kinds of memory maps: 1) the initial target scribble map $\bm M_{s}\!\in\! \{0,1\}^{1\times H\times W}$, which is the ground truth given by the user; 2) the predicted probability maps $\bm M_{r}\!\in\! (0,1)^{T\times H\times W}$ for segmenting the region that is highly similar to the initial target scribble region in previous frames (we refer to such regions as \emph{reliable regions} since they are with high confidence belonging to the target); 3) the predicted probability maps $\bm M_{e}\!\in\! (0,1)^{T\times H\times W}$ for segmenting the entire target in previous frames. Here, $T$ is the number of memory frames. As the corresponding covered areas dilate from the scribble to the reliable region and further to the entire target, the reliability of the memory map decreases.

With such memory, RHMNet processes a new query frame in two steps: 1) capture the reliable region in the query frame according to the initial target scribble memory and the reliable region memory; 2) expand the newly captured reliable region to the entire target conditioned on the captured reliable region itself and all previous memory information.

\vspace{1mm}
\subsubsection{Capturing the reliable region}
To capture the reliable region in a new query frame, RHMNet first encodes the historical target information stored in $\bm M_{s}$ and $\bm M_{r}$ into the features of the memory frames $\bm F^{m}\!\in\! \mathbb{R}^{T\times h\times w\times c}$, yielding the encoded memory features $\bm E_{s}^{m}\!\in\! \mathbb{R}^{1\times h\times w\times c}$ and $\bm E_{r}^{m}\!\in\! \mathbb{R}^{T\times h\times w\times c}$, respectively. We refer to this operation as memory encoding. Then, RHMNet matches the feature of the query frame $\bm F^{q}\in \mathbb{R}^{1\times h\times w\times c}$ with these encoded memory features to obtain the query representation $\bm R_{r}^{q}\in \mathbb{R}^{1\times h\times w\times c}$ that models the reliable region information in the query frame. After that, a segmentation head is built on $\bm R_{r}^{q}$ to segment the reliable region.
Denoting the matching module and segmentation head for the reliable region as ${\rm \Phi}_{r}^{\rm Match}$ and ${\rm \Phi}_{r}^{\rm Head}$, respectively, the predicted probability map $\bm Q_{r}\in (0,1)^{1\times H \times W}$ for the reliable region in the query frame is:
\begin{equation}
\label{Eq:tracking}
\begin{split}
    & \bm R_{r}^{q} = {\rm \Phi}_{r}^{\rm Match}(\bm E_{s}^{m} \uplus \bm E_{r}^{m}, \bm F^{q}),\\
    & \bm Q_{r} = {\rm \Phi}_{r}^{\rm Head}(\bm R_{r}^{q}, \bm f^{q}).
\end{split}
\vspace{-1mm}
\end{equation}
Here, $\uplus$ denotes the concatenation operation in the time dimension, and $\bm f^{q}$ refers to the intermediate-layer features of the query frame that provide more spatial details. Note that this step does not involve $\bm M_{e}$ as it is less reliable and may disturb the prediction of the reliable region.
\begin{figure*}[t]
\centering
    \includegraphics[width=1\textwidth]{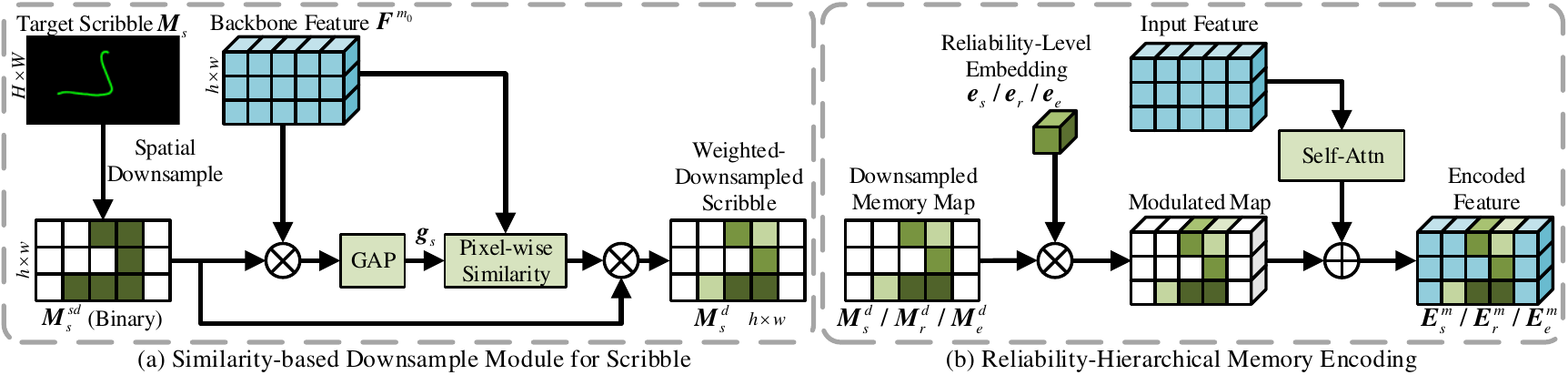}
    \caption{(a) Illustration of the similarity-based downsample module. It performs the weighted downsample by calculating the pixel-wise similarity between the feature of the initial frame and the global representation of the target scribble region. (b) Illustration of the memory encoding process. The reliability-level embeddings, $\bm e_{s}$, $\bm e_{r}$, and $\bm e_{e}$ which are different learnable vectors, are used to differentiate the degree of the reliability of target information in different memory maps. $\otimes$ and $\oplus$ denote element-wise multiplication and element-wise summation, respectively.}
\label{Fig:Encoding}
\vspace{-4mm}
\end{figure*}

\vspace{1mm}
\subsubsection{Expanding to the entire target}
To expand to the entire target in the query frame, we take the newly captured reliable region and all previous memory information as the reference. Therefore, we encode the reliable region information in $\bm Q_{r}$ into the feature of the query frame $\bm F^{q}$, obtaining the encoded query feature $\bm E^{q}_{r} \in \mathbb{R}^{1\times h\times w\times c}$. On the other hand, we encode the historical target information stored in $\bm M_{e}$ into $\bm E_{r}^{m}$ to obtain newly encoded memory features $\bm E_{e}^{m}$. In this way, $\bm E_{e}^{m}$ model the historical target information in both $\bm M_{r}$ and $\bm M_{e}$. After that, similar to the implementation for capturing the reliable regions, a matching module and a segmentation head are constructed to segment the entire target.
Denoting the matching module and the segmentation head for the entire target as ${\rm \Phi}^{\rm Match}_{e}$ and ${\rm \Phi}^{\rm Head}_{e}$, respectively, the predicted probability map $\bm Q_{e}\in (0,1)^{1\times H \times W}$ for the entire target in the query frame is:
\begin{equation}
\label{Eq:tracking}
\begin{split}
    & \bm R_{e}^{q} = {\rm \Phi}_{e}^{\rm Match}(\bm E_{s}^{m} \uplus \bm E_{e}^{m} \uplus \bm E^{q}_{r}, \bm R^{q}_{r}),\\
    & \bm Q_{e} = {\rm \Phi}_{e}^{\rm Head}(\bm R_{e}^{q}, \bm f^{q}).
\end{split}
\vspace{-1mm}
\end{equation}
Herein $\bm R_{e}^{q}$ denotes the query representation outputted by the matching module, which models the entire target information in the query frame.

\subsection{Memory encoding}
\label{Sec: Memory Encoding}
Memory encoding aims to encode the reliability-hierarchical memory information into the features of the memory frames. To encode the reliability level of the three types of memory maps $\bm M_{s}$, $\bm M_{r}$, and $\bm M_e$, we introduce three reliability-level embeddings, $\bm e_{s}$, $\bm e_{r}$, and $\bm e_e$ to represent their reliability levels. Technically, these embeddings are different \emph{learnable vectors}. Considering that the spatial size of the extracted feature is 1/16 of that of the memory maps, we downsample the memory maps spatially before performing memory encoding.

\vspace{1mm}
\subsubsection{Downsampling}
For the numerically continuous probability maps of the reliable region and the entire target, directly downsampling them via interpolation almost does not lose the precision. However, for the binary target scribble map, directly interpolating it brings unbearable spatial quantization errors.

\begin{figure}[t]
\centering
    \includegraphics[width=1.0\columnwidth]{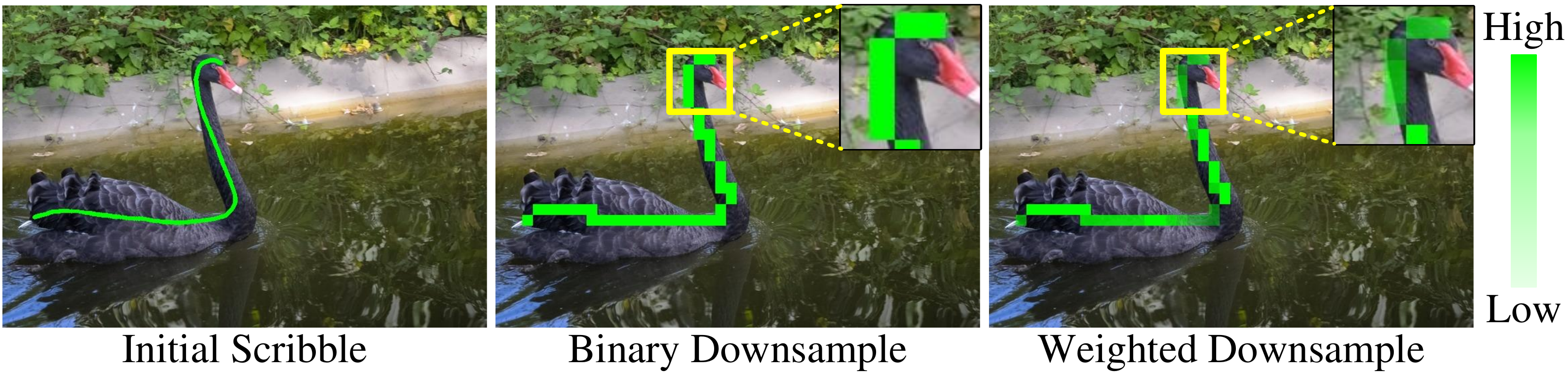}
    \vspace{-2mm}
    \caption{Comparison of the binary downsampled scribble and the weighted downsampled one. Unlike the binary downsample method, the proposed similarity-based downsample approach adaptively assigns a lower value for the patch in which the target scribble and the background co-exist.}
\label{Fig:Visualization_Scribble}
\end{figure}

To handle this issue, we propose a similarity-based downsample module for the target scribble map, as shown in Figure~\ref{Fig:Encoding} (a). Specifically, we first divide the target scribble map $\bm M_s$ into non-overlapping patches of $16\times 16$. Then we assign a binary value for every patch according to whether the target scribble passes through the patch to get a binary downsampled target scribble map $\bm M_{s}^{bd}\in\{0,1\}^{1\times h \times w}$. After that, we filter the feature of the initial frame $\bm F^{m_{0}}$ with $\bm M_{s}^{bd}$, and perform Global Average Pooling (GAP) on the filtered feature to obtain the global representation $\bm g_{s}$ of the target scribble region. Finally, we compute the pixel-wise cosine similarity between $\bm g_{s}$ and $\bm F^{m_{0}}$, and filter the resulting similarity map with $\bm M_{s}^{bd}$ to obtain the weighted downsampled scribble $\bm M_{s}^{d}$:
\begin{equation}
\label{Eq:tracking}
\begin{split}
    & \bm g_{s} = \phi^{\rm GAP}(\bm F^{m_{0}}\otimes \bm M_{s}^{bd}),\\
    & \bm M_{s}^{d} = \bm M_{s}^{bd} \otimes \phi^{\rm Sim}(\bm g_{s}, \bm F^{m_{0}}),
\end{split}
\end{equation}
where $\phi^{\rm GAP}$ and $\phi^{\rm Sim}$ refer to the GAP and cosine similarity operations, respectively. $\otimes$ denotes the element-wise multiplication operation. 
As shown in Figure~\ref{Fig:Visualization_Scribble}, the patch in which the target scribble and the background co-exist is assigned a lower value on the weighted downsampled target scribble map, alleviating the effect of spatial quantization errors.

\vspace{1mm}
\subsubsection{Process of memory encoding} 
As shown in Figure~\ref{Fig:Encoding} (b), the memory encoding processes for different memory maps are the same, except for the used reliability-level embedding.
Without loss of generality, we present this process taking the target scribble encoding as an example. Given the reliability-level embedding $\bm e_{s}$ for the target scribble, we first incorporate the reliability-level information into the downsampled target scribble map $\bm M_{s}^{d}$. To this end, we calculate the element-wise multiplication of $\bm e_{s}$ and $\bm M_{s}^{d}$, yielding a modulated map. Meanwhile, we use a self-attention~\cite{Transformer} layer to enhance $\bm F^{m_{0}}$ via modeling the relation between feature pixels.
Finally, we combine the modulated map and the enhanced feature to obtain the encoded feature $\bm E_{s}^{m}$ modeling the information represented by the target scribble. This process can be formulated as:
\begin{equation}
\label{Eq:encoding}
    \bm E_{s}^{m} = (\bm M_{s}^{d} \otimes \bm e_{s}) \oplus {\rm \Phi}^{sa}_{enc}(\bm F^{m_0}),
\end{equation}
where ${\rm \Phi}^{sa}_{enc}$ refers to the self-attention layer in the memory encoding module.

\subsection{Memory matching}
Taking the memory information as the reference, the matching module aims to decode the target state in the query image via pixel-level feature matching to obtain query representations for further segmentation. To this end, we resort to the transformer decoder~\cite{Transformer} architecture to build our matching module, which mainly consists of a self-attention layer and a cross-attention layer. Next, we detail the matching module taking the one for reliable regions as an example. Herein the self-attention layer is used to enhance the feature of the query frame $\bm F^{q}$ by modeling the relation between the feature pixels. Then the cross-attention layer takes as input the encoded feature of the memory frames $\bm E_{s}^{m} \uplus \bm E_{r}^{m}$ (serving as the key and value in cross-attention) and the enhanced feature of the query frames $\bm F^{q}_{e}$ (serving as the query in cross-attention), and outputs the desired query representation. Thus, the matching operation ${\rm \Phi}_{r}^{\rm Match}$ for segmenting the reliable region can be formulated as:
\begin{equation}
\label{Eq:tracking}
\begin{split}
    {\rm \Phi}_{r}^{\rm Match}(\bm E_{s}^{m} \uplus \bm E_{r}^{m}, \bm F^{q})={\rm \Phi}^{ca}_{r}({\rm \Phi}^{sa}_{r}(\bm F^{q}_{e}), \bm E_{s}^{m} \uplus \bm E_{r}^{m}),
\end{split}
\vspace{-1mm}
\end{equation}
where ${\rm \Phi}^{sa}_{r}$ and ${\rm \Phi}^{ca}_{r}$ denote the self-attention and cross-attention layers in the matching module for segmenting the reliable region, respectively. The matching operation for the entire target ${\rm \Phi}^{\rm Match}_{e}$ shares the same process with ${\rm \Phi}^{\rm Match}_{r}$. Therefore, we omit its formulation for clarity.

\subsection{Scribble-supervised learning}
\subsubsection{Scribble-supervised learning for reliable region segmentation}
The reliable region in a query frame is defined as the region highly similar to the target scribble region in the initial frame. Although the target appearance varies in different frames, the target scribble regions, \ie~skeleton regions, in multiple frames are generally similar to each other. Thus we could use the target scribble as the supervision for learning to segment the reliable region. 
To this end, we define the pixels on the target scribble as positive ones. For those out of the target scribble, we divide them into two types: the pixels nearby the target scribble are defined to be ignored pixels, and the remaining pixels are defined as negative pixels. The rationale behind the definition is that the pixels close to the target scribble are usually similar to the target scribble region, and thus defining them as negative pixels would affect the convergence of the model. Considering the imbalance between the positive and negative pixels, we opt for the focal loss~\cite{FocalLoss}. Thus, The loss function for the probability map of the reliable region over a sequence training sample is:
\vspace{-1mm}
\begin{equation}
\label{Eq:focalloss}
\begin{split}
    \mathcal{L}^{FL}_{r}=\frac{1}{N}\sum_{t=0}^{L-1}\sum_{(i,j)\in s^{t} }\phi^{fl}(\bm Q_{r}^{t,(i,j)}, \bm Y^{t,(i,j)}_{r}).
\end{split}
\vspace{-1mm}
\end{equation}
Here, $\mathcal{\phi}^{fl}$ is the focal loss function. $\bm Q_{r}^{t}$ and $\bm Y_{r}^{r}$ refer to the probability map of the reliable region and the ground truth on the $t$-th frame, respectively. $(i,j)$ denotes the pixel index. $s^{t}$ is the index set of the pixels except for the ignored ones in the $t$-th frame. $N$ is the total pixel number except for the ignored ones. $L$ is the length of the sequence training sample.

\vspace{1mm}
\subsubsection{Scribble-supervised learning for entire target segmentation}
Following WS-Seg~\cite{NormalizedCut-WSSS}, we use the Partial Cross-Entropy (PCE) loss~\cite{NormalizedCut-WSSS} to exploit the scribble annotation for model learning, which only imposes sparse supervision. To obtain dense supervision, we further introduce a smoothness loss and a bidirectional prediction consistency loss.

\vspace{1mm}
\noindent\textbf{Partial cross-entropy loss.}~The PCE loss calculates the cross-entropy only in the area where the scribble annotation passes while ignoring other regions. Since our method processes each object independently, we first aggregate the independent probability map $\bm Q_{e}^{t}$ of every object to obtain a multi-object probability map $\bm Q_{e,mo}^{t}$. Thus, the PCE loss is formulated as:
\vspace{-1mm}
\begin{equation}
\label{Eq:pceloss}
    \mathcal{L}^{PCE}_{e}=-\frac{1}{N'}\sum_{t=0}^{L-1}\sum_{\substack{(i,j)\in\Omega_{t}}}log(\bm Q_{e,mo}^{t,(i,j),c}),
\vspace{-1mm}
\end{equation}
where $\bm Q_{e,mo}^{t,(i,j),c}$ denotes the probability that the pixel located at $(i,j)$ in the $t^{th}$ frame is predicted as its ground-truth object identity $c$. 
$\Omega_{t}$ is the index set of the labeled pixels in the $t$-th frame. 
$N^{'}$ is the total number of the labeled pixels.

\vspace{1mm}
\noindent\textbf{Smoothness loss.} To provide dense supervision for the predicted mask, we introduce the smoothness loss to utilize the pair-wise pixel relations in the image, inspired by~\cite{UMDE}. The intuition is that two adjacent pixels should have an approximately equal probability of belonging to the target object if they are similar to each other. Specifically, the smoothness loss imposes an adaptive smoothing constraint on the predicted probability map according to the image smoothness. Formally, the smoothness loss on a sequence training sample is:
\vspace{-1mm}
\begin{equation}
\label{Eq:smoothloss}
\begin{split}
    \mathcal{L}^{SN}_{e}=\frac{1}{N''}\sum_{t=0}^{L-1}\sum_{d\in \vec{x},\vec{y}}\sum_{(i,j)}(\partial_{d}\bm Q_{e}^{t,(i,j)} \cdot e^{-\alpha \lVert\partial_{d} \bm I^{t,(i,j)} \rVert}).
\end{split}
\vspace{-1mm}
\end{equation}
Here, $\vec{x}$ and $\vec{y}$ refer to the partial derivatives in the horizontal and vertical directions, respectively. $\bm Q_{e}^{t}$ is the predicted probability map of the entire target on the $t$-th frame. $\bm I_{t}$ is the image intensity of the $t$-th frame. $N''$ is the total pixel number. $\alpha$ is a controlling factor for the smoothing constraint.

\vspace{1mm}
\noindent\textbf{Bidirectional prediction consistency loss.}
In a sequence training sample, the target appearance usually varies across frames. Thus performing VOS in the forward-prediction way (from the first frame to the last one) or the backward-prediction way (from the last frame to the first one) will obtain different segmentation results.
Based on this intuition, we propose a Bidirectional Prediction Consistency (BPC) loss to exploit the variations of the target appearance across different frames for weak-supervised learning. By imposing the consistency constraint between the forward predictions and backward predictions, the BPC loss excavates and exploits the frame-level relation within a sequence training sample. Formally, the BPC loss over a sequence training sample is defined as:
\vspace{-1mm}
\begin{equation}
\label{Eq:smoothloss}
\begin{split}
    \mathcal{L}^{BPC}_{e}=\frac{1}{N''}\sum_{t=0}^{L-1}\sum_{(i,j)}\lVert \bm Q_{e}^{t,(i,j)} - \bm B_{e}^{t,(i,j)}\rVert^{2},
\end{split}
\vspace{-1mm}
\end{equation}
where $\bm Q_{e}^{t}$ and $\bm B_{e}^{t}$ are the computed probability maps on the $t$-th frame in forward and backward predictions, respectively.

\vspace{1mm}
\subsubsection{Overall loss function}
The whole model of our RHMNet can be trained using sequence training samples with scribble annotations end-to-end. The parameters are optimized by integrating all the above-mentioned loss functions:
\vspace{-1mm}
\begin{equation}
\label{Eq:smoothloss}
\begin{split}
    \mathcal{L}=\lambda_{1}\mathcal{L}^{FL}_{r}+\lambda_{2}\mathcal{L}_{e}^{PCE}+\lambda_{3}\mathcal{L}^{SN}_{e}+\lambda_{4}\mathcal{L}^{BPC}_{e},
\end{split}
\vspace{-1mm}
\end{equation}
where $\lambda_{1}$, $\lambda_{2}$, $\lambda_{3}$, and $\lambda_{4}$ are hyperparameter weights to balance the loss functions. In our implementation, they are tuned to be 1, 1, 0.3, and 20, respectively.

\begin{algorithm}[t]
    \small
    \begin{spacing}{1.5}
    \caption{Inference with our proposed RHMNet}
    \label{vos_algor}
        \LinesNumbered
        \KwIn{The testing video $\{\bm I^{t}\}_{t=0}^{T-1}$, the initial scribble $\bm M_{s}$}
        \KwOut{Prediction results $\{\bm Q^{t}_{e}\}^{T-1}_{t=1}$}
        \For{$t=0$ \KwTo $T-1$}
        {
            $\bm f^{t}, \bm F^{t} \gets \text{Extractor}(\bm I^{t})$\;
            \If(\tcp*[h]{\scriptsize{Initialize the memory.}}){$t=0$}
            {
            $\bm E^{m}_{s}, \bm E^{m}_{r}, \bm E^{m}_{e} \gets \text{Encode}(\bm F^{0}, \bm M_{s}), \varnothing, \varnothing$\;
            }
            $\bm R_{r}^{t} \gets {\rm \Phi}_{r}^{\rm Match}(\bm E_{s}^{m} \uplus \bm E_{r}^{m}, \bm F^{t})$\;
            $\bm Q_{r}^{t} \gets {\rm \Phi}_{r}^{\rm Head}(\bm R_{r}^{t}, \bm f^{t})$~\tcp{\scriptsize{Predict the probability map for the reliable region}}
            $\bm E_{r}^{q} \gets \text{Encode}(\bm Q_{r}^{t}, \bm F^{t})$\;
            $\bm R_{e}^{t} \gets {\rm \Phi}_{e}^{\rm Match}(\bm E_{s}^{m} \uplus \bm E_{e}^{m} \uplus \bm E_{r}^{q}, \bm R_{r}^{t})$\;
            $\bm Q_{e}^{t} \gets {\rm \Phi}_{r}^{\rm Head}(\bm R_{e}^{t}, \bm f^{t})$~\tcp{\scriptsize{Predict the probability map for the entire target}}
            \If(\tcp*[h]{\scriptsize{Update the memory.}}){$t~{\rm mod}~6 = 0$}
            {
                $\bm E^{m}_{r}, \bm E^{m}_{e} \gets \text{Update}(\bm E^{m}_{r}, \bm E^{m}_{e}, \bm Q_{r}^{t}, \bm Q_{e}^{t}, \bm F^{t})$\;
            }
        }
        \end{spacing}
\end{algorithm}

\subsection{Inference}
Following many VOS methods~\cite{STM-VOS,KMN-VOS,HMMNet-VOS}, we process each object independently to predict the probability map and then merge these maps with a soft aggregation operation~\cite{STM-VOS} when encountering multiple target objects. Algorithm~\ref{vos_algor} summarizes the inference process for one single object with our RHMNet and omits the process for multiple objects for presentation clarity.
In practical implementation, we directly manage the encoded memory features $\bm E_{s}^{m}$, $\bm E_{r}^{m}$, and $\bm E_{e}^{m}$ to maintain the memory information. In the first frame, the encoded memory features $\bm E_{s}^{m}$ is initialized with $\bm F^{0}$ and $\bm M_{s}$, and $\bm E_{r}^{m}$, and $\bm E_{e}^{m}$ are initialized to empty sets. With the reliability-hierarchical memory, RHMNet processes every testing frame, \ie~the query frame, in two steps as described in Section~\ref{Sec: Framework}: 1) segmenting the reliable region based on the memory information with high reliability as shown in line 6 and line 7, and 2) then expanding it to the entire target by taking all the memory information as the reference as shown from line 8 to line 10. After predicting the probability map $\bm Q_{e}^{t}$ for each object, RHMNet merges these independently predicted maps together and then accordingly generates the segmentation mask for all the objects via \emph{argmax}. To maintain the reliability-hierarchical memory, we update $\bm E_{r}^{m}$ and $\bm E_{e}^{m}$ every six frames with the newly predicted probability maps $\bm Q_{r}^{t}$ and $\bm Q_{e}^{t}$ and the new backbone feature $\bm F^{t}$. The encoded memory features from the earliest frame will be removed except for those from the initial one when reaching the pre-defined memory capacity $M_{c}$.

\section{Experiments}
\newcommand{\tabincell}[2]{\begin{tabular}{@{}#1@{}}#2\end{tabular}}
\begin{table*}[t]
\setlength\tabcolsep{6.25pt}
\begin{center}
\caption{Experimental results of six variants of our RHMNet on the DAVIS 2017 validation set. Youtube-VOS is additionally used for training these variants. The reported scores are the average results of using five groups of the initial scribbles for testing. The best scores are denoted in bold font.}
\renewcommand\arraystretch{1.0}
\resizebox{0.875\linewidth}{!}{
\begin{tabular}{ccc|cccc|ccc}
\toprule
           \tabincell{c}{Reliable-Hierarchical\\Mechanism} &
           \tabincell{c}{Step-wise\\Expanding} &
           \tabincell{c}{Similarity-based\\Downsample} &
           \tabincell{c}{$\mathcal{L}_{e}^{PCE}$} &
           \tabincell{c}{$\mathcal{L}_{e}^{SN}$} &
           \tabincell{c}{$\mathcal{L}_{e}^{BPC}$} &
           \tabincell{c}{$\mathcal{L}_{r}^{FL}$} &
           \tabincell{c}{$\mathcal{J} \& \mathcal{F}$} &
           \tabincell{c}{$\mathcal{J}$} &
           \tabincell{c}{$\mathcal{F}$}\\
\midrule
\checkmark & \checkmark & \checkmark & \checkmark & \checkmark & \checkmark & \checkmark & \textbf{67.4} & \textbf{66.2} & \textbf{68.6} \\
$\times$ & \checkmark & \checkmark & \checkmark & \checkmark & \checkmark & \checkmark & 64.1 & 62.6 & 65.5 \\
\checkmark & $\times$ & \checkmark & \checkmark & \checkmark & \checkmark & $\times$ & 63.8 & 62.4 & 65.4 \\
\checkmark & \checkmark & $\times$ & \checkmark & \checkmark & \checkmark & \checkmark & 65.8 & 64.0 & 67.7 \\
\midrule
\checkmark & \checkmark & \checkmark & \checkmark & \checkmark & $\times$ & \checkmark & 65.6 & 64.4 & 66.8 \\
\checkmark & \checkmark & \checkmark & \checkmark & $\times$ & $\times$ & \checkmark & 61.0 & 60.1 & 61.9 \\
\bottomrule
\end{tabular}}
\label{Tab:Ablation}
\end{center}
\end{table*}

\subsection{Implementation details} 
\vspace{1mm}
\subsubsection{Model details} We use ResNet-50~\cite{ResNet} pre-trained on ImageNet~\cite{Imagenet} as the backbone and the output of \emph{conv4} as the backbone feature. 
The segmentation head in our model consists of three refinement modules with skip-connection to the intermediate layers of the backbone, which gradually increases the spatial size of the query representation from $\frac{H}{16}\times \frac{W}{16}$ to $\frac{H}{4}\times \frac{W}{4}$. Based on the refined query representation, we adopt a convolutional layer followed by an interpolation operation and a softmax layer to predict the final probability map with a size of $H\times W$.

\vspace{1mm}
\subsubsection{Training details} 
Similar to many approaches~\cite{STM-VOS,KMN-VOS,HMMNet-VOS}, we adopt a two-stage training scheme: training our model first on static images and then on video sequences. In the first stage, COCO~\cite{COCO} is used, and the sequence training sample is generated by augmenting the static image. 
In the second stage, DAVIS~\cite{DAVIS} and Youtube-VOS~\cite{Youtube-vos} are used. As a proof of concept for the proposed approach, we synthesize the object scribble annotations and background scribble annotations according to the ground-truth instance-level mask annotations. We use the method proposed in~\cite{UFO2} to synthesize the scribble annotations. Following STM~\cite{STM-VOS}, we use the dynamic memory strategy in the two training stages.
The loss function $\mathcal{L}_{e}^{BPC}$, which involves bidirectional prediction, is only used in the second stage, as the target appearance remains unchanged in the first stage. To reduce the GPU memory usage, we randomly stop the gradient propagation of either the forward or the backward predictions.
The sequence training sample length $L$ is 3.
AdamW~\cite{AdamW} with a weight decay of $10^{-4}$ is used to optimize our model. For the first training stage, we train our model for 14 epochs. The learning rate is set to $5 \times 10^{-5}$ and drops by a factor of 2 every 6 epochs. In the second stage, we train our model for 80 epochs. The learning rate is set to $ 10^{-4}$ and drops in a polynomial manner.

\vspace{1mm}
\subsubsection{Inference details}
The maximum memory capacity $M_{c}$ is set to four according to the ablation study results. We test our method on an RTX3090 GPU, and the average running speed on DAVIS 2017~\cite{DAVIS} is about 18.4 Frames Per Second~(FPS). Note that we do not need the users to annotate the background scribble in the first frame for inference. Our project is available at \emph{https://github.com/mkg1204/RHMNet-for-SSVOS}.

\subsection{Benchmarks and metrics} 
DAVIS~\cite{DAVIS} is a popular VOS benchmark including two versions: DAVIS 2016 for single-object segmentation and DAVIS 2017 for multi-object segmentation.
DAVIS uses region similarity $\mathcal{J}$ and contour accuracy $\mathcal{F}$ to evaluate the performance of VOS algorithms. Herein $\mathcal{J}$ and $\mathcal{F}$ represent the average IoU and the boundary similarity between the predicted masks and the ground truth annotations, respectively. The average of them, $\mathcal{J} \& \mathcal{F}$, is used to evaluate the overall performance.
Youtube-VOS~\cite{Youtube-vos} is a large-scale dataset for multi-object segmentation which contains 26 unseen categories in the validation set.
Thus, it calculates region similarity ($\mathcal{J}_\mathcal{S}$, $\mathcal{J}_\mathcal{U}$) and contour accuracy ($\mathcal{F}_\mathcal{S}$, $\mathcal{F}_\mathcal{U}$) on the seen set and unseen set separately. The average score $\mathcal{G}$ is used to assess the overall quality of the segmentation results.

To evaluate our model on DAVIS and Youtube-VOS, we manually annotate the scribble for every target when it first appears in the video.
The principle for manually annotating is to draw the scribble as simply as possible on the premise that it can coarsely reflect the skeleton of the target. Particularly, we invite five users to annotate five groups of initial scribbles on the DAVIS 2017 validation set for further ablation study. We also measure the average time cost for annotating the scribble using a pencil, which is 7.0 seconds per object. 

\subsection{Ablation studies}
We conduct experiments to investigate the effect of each proposed component, the robustness to the randomness of the initial scribbles, and the influence of the memory capacity.

\begin{table}[t]
    \begin{center}
    \small
    \setlength\tabcolsep{2.5pt}
    \caption{Experimental results of using five groups of initial scribbles for testing on Davis 2017 validation set.}
    \label{Tab:VOS-Robust_to_scribble}
    \renewcommand\arraystretch{1.05}
    \begin{tabular}{cccccccc}
    \toprule
            &
        Group1   & 
        Group2   & 
        Group3   & 
        Group4   & 
        Group5   & 
        Range    &
        Std.Dev.\\
    \midrule
    $\mathcal{J} \& \mathcal{F}$ & 67.8 & 67.1 & 66.9 & 67.0 & 68.2 & 1.3 & 0.51\\
    $\mathcal{J}$ & 66.8 & 66.0 & 65.8 & 65.6 & 67.0 & 1.4 & 0.56\\
    $\mathcal{F}$ & 68.8 & 68.2 & 68.0 & 68.5 & 69.5 & 1.5 & 0.53\\
    \bottomrule
    \end{tabular}
    \end{center}
\end{table}

\begin{table}[t]
    \begin{center}
    \small
    \setlength\tabcolsep{6.5pt}
    \caption{Experimental results and running speed of our RHMNet under different memory capacity on DAVIS 2017 validation set. The reported scores are the average results of using five groups of the initial scribbles for testing.}
    \label{Tab:VOS-Mem_size}
    \renewcommand\arraystretch{1.05}
    \begin{tabular}{cccccc}
    \toprule
              &
        $M_{c}=1$ & 
        $M_{c}=2$ & 
        $M_{c}=4$ & 
        $M_{c}=8$ &
        $M_{c}=16$ \\
    \midrule
    $\mathcal{J} \& \mathcal{F}$ & 64.8 & 66.6 & \textbf{67.4} & \underline{67.2} & 67.1 \\
    $\mathcal{J}$ & 63.5 & 65.3 & \textbf{66.2} & \underline{66.1} & 66.0 \\
    $\mathcal{F}$ & 66.3 & 67.9 & \textbf{68.6} & \underline{68.4} & 68.3 \\
    \midrule
    $\text{FPS}$ & \textbf{21.7} & \underline{20.2} & 18.4 & 17.1 & 16.6 \\
    \bottomrule
    \end{tabular}
    \end{center}
\end{table}

\vspace{1mm}
\subsubsection{Effect of different component}
To investigate the effect of each proposed component, we perform ablation experiments over six variants of our RHMNet on DAVIS 2017, as shown in Table~\ref{Tab:Ablation}. Note that we report the average result of using five groups of the initial scribbles for testing.

\vspace{1mm}
\noindent\textbf{Effect of the reliability-hierarchical mechanism.}
To investigate the effect of the reliability-hierarchical mechanism, we replace the three reliability-level embeddings $\bm e_{s}$, $\bm e_{r}$, and $\bm e_{e}$ with the same target embedding. Thus, this variant (the second row) is agnostic of the reliability of $\bm M_s$, $\bm M_r$, and $\bm M_e$. Compared with our intact model (the first row), removing the reliability-hierarchical mechanism leads to a performance drop of 3.6\% in $\mathcal{J} \& \mathcal{F}$, which manifests the effectiveness of the proposed reliability-hierarchical mechanism.
 
\vspace{1mm}
\noindent\textbf{Effect of the step-wise expanding strategy.}
To analyze the effect of the step-wise expanding strategy, we skip the reliable region prediction step and directly predict the entire target mask in a single step (the third row). The performance gap between this variant and our intact model validates the effectiveness of the step-wise expanding strategy, which benefits alleviating the error accumulation issue.

\vspace{1mm}
\noindent\textbf{Effect of the similarity-based downsample module.} 
To study the effect of the similarity-based downsample module, we remove it from our approach and direct use the binary downsampled scribble map (the fourth row). The binary downsampled scribble map brings large spatial quantization errors and consequently results in a non-negligible performance drop. The performance gap shows that our proposed similarity-based downsample method can effectively alleviate the adverse effect of the spatial downsample.

\vspace{1mm}
\noindent\textbf{Effect of the bidirectional prediction consistency loss.} 
To analyze the effect of the BPC loss, we remove $\mathcal{L}_{e}^{BPC}$ from the total loss function $\mathcal{L}$ and use the remaining loss functions to train the proposed model (the fifth row). The performance drop manifests that $\mathcal{L}_{e}^{BPC}$ benefits the scribble-supervised learning for video object segmentation.

\vspace{1mm}
\noindent\textbf{Effect of the smoothness loss.} 
In the last row of Table~\ref{Tab:Ablation}, we further remove the smoothness loss $\mathcal{L}_{e}^{SN}$ from the total loss function $\mathcal{L}$, which leads to a considerable performance drop of 4.6\%. It demonstrates that dense supervision for predicting the entire target mask is crucial, and the pair-wise pixel relation is an effective cue for dense supervision.

\vspace{1mm}
\subsubsection{Robustness to the randomness of the initial scribbles.} 
To evaluate the robustness to the randomness of the initial scribbles, we report the experimental results of using five groups of initial scribbles annotated by different users on DAVIS 2017 in Table~\ref{Tab:VOS-Robust_to_scribble}. The results of initializing with different scribbles are stable with a standard deviation of 0.51 in $\mathcal{J} \& \mathcal{F}$, demonstrating that our model is robust to the randomness of the initial scribbles.

\vspace{1mm}
\subsubsection{Effect of the memory capacity.}
To investigate the effect of the memory capacity $M_c$, we measure the performance of our method with different memory capacities and report the experimental results in Table~\ref{Tab:VOS-Mem_size}. The performance improves along with the memory capacity $M_c$ increasing and saturates at about $M_c=4$. On the other hand, the running speed decreases along with the memory capacity $M_c$ increasing.

\begin{table}[t]
\setlength\tabcolsep{1.5pt}
\begin{center}
\caption{Experimental results on the validation set of DAVIS. $^{\dagger}$ denotes using external training datasets. `None' denotes the methods are trained in a self-supervised manner. The scores of the scribble-initialized methods except for~\cite{SSVOS} are the average results of using five groups of initial scribbles for testing. $*$~denotes that the scribbles are generated by the authors.}
\renewcommand\arraystretch{1.0}
\resizebox{1\linewidth}{!}{
\begin{tabular}{l|c|c|ccc|ccc}
\toprule
\multirow{2}{*}{Methods} & Init. & Training & \multicolumn{3}{c}{DAVIS 2016} \vline & \multicolumn{3}{c}{DAVIS 2017} \\
 & Anno. & Label & $\mathcal{J} \& \mathcal{F}$ & $\mathcal{J}$ & $\mathcal{F}$ & $\mathcal{J} \& \mathcal{F}$ & $\mathcal{J}$ & $\mathcal{F}$ \\
\midrule
FEEL$^{\dagger}$~\cite{FEEL-VOS} & Mask & Mask & 81.7 & 81.1 & 82.2 & 71.6 & 69.1 & 74.0 \\
RGMP~\cite{RGMP-VOS}                   & Mask & Mask & 81.8 & 81.5 & 82.0 & 66.7 & 64.8 & 68.6 \\
A-GAME$^{\dagger}$~\cite{GAM-VOS}   & Mask & Mask & 82.1 & 82.0 & 82.2 & 70.0 & 67.2 & 72.7 \\
LWL$^{\dagger}$~\cite{LWL-VOS}      & Mask & Mask & --   & --   & --   & 81.6 & 79.1 & 84.1 \\
STM$^{\dagger}$~\cite{STM-VOS}      & Mask & Mask & 89.3 & 88.7 & 89.9 & 81.8 & 79.2 & 84.3 \\
CFBI$^{\dagger}$~\cite{CFBI-VOS}    & Mask & Mask & 89.4 & 88.3 & 90.5 & 81.9 & 79.1 & 84.6 \\
HMMNet$^{\dagger}$~\cite{HMMNet-VOS}& Mask & Mask & \textbf{90.8} & \textbf{89.6} & \textbf{92.0} & \textbf{84.7} & \textbf{81.9} & \textbf{87.5} \\
\midrule
SiamMask$^{\dagger}$~\cite{SiamMask}& Box  & Mask & 69.8 & 71.7 & 67.8 & 56.4 & 54.3 & 58.5 \\
E3SN$^{\dagger}$~\cite{E3SN}        & Box  & Mask & 71.2 & 73.0 & 69.3 & 58.0 & 56.1 & 59.8 \\
SRCNN$^{\dagger}$~\cite{SiamRCNN}& Box  & Mask & 79.7 & 78.1 & 81.2 & 70.6 & 66.1 & 75.0 \\
LWL-Box$^{\dagger}$~\cite{LWL-VOS}  & Box  & Mask & --   & --   & --   & 70.6 & 67.9 & 73.3 \\
QMA$^{\dagger}$~\cite{QMRA-VOS}     & Box  & Mask & \textbf{85.9} & \textbf{84.7} & \textbf{87.1} & \textbf{71.9} & --   & --   \\
\midrule
Vid.Color.~\cite{VidColor}             & Mask & None & --   & --   & --   & 33.7 & 34.6 & 32.7 \\
CycT~\cite{CycleTime}             & Mask & None & --   & --   & --   & 48.7 & 46.4 & 50.0 \\
UVC~\cite{uvc}                         & Mask & None & --   & --   & --   & 58.9 & 57.7 & 60.0 \\
MAST~\cite{MAST-VOS}                   & Mask & None & --   & --   & --   & \textbf{65.5} & \textbf{63.3} & \textbf{67.6} \\
\midrule
MAST-Box            & Box & Mask & 62.7 & 59.5 & 65.8 & 55.1 & 52.2 & 57.9 \\
MAST-Scribble            & Scribble & None & 45.4 & 43.8 & 46.9 & 40.3 & 38.6 & 42.0 \\
SS-STM$^{\dagger}$     & Scribble & Scribble & 73.5 & 71.8 & 74.4 & 53.3 & 51.4 & 55.1 \\
Huang \etal$^{\dagger}$~\cite{SSVOS}    & Scribble$^*$ & Scribble$^*$ & -- & -- & -- & 55.7 & 53.4 & 57.9 \\
SS-HMMNet$^{\dagger}$     & Scribble & Scribble & 78.7 & 78.0 & 79.5 & 59.5 & 58.4 & 60.6 \\
\textbf{Ours}$^{\dagger}$                    & Scribble & Scribble & \textbf{82.2} & \textbf{81.3} & \textbf{83.0} & \textbf{67.4} & \textbf{66.2} & \textbf{68.6} \\
\bottomrule
\end{tabular}}
\label{Tab:DAVIS}
\end{center}
\end{table}

\subsection{Comparisons with other methods}

\vspace{1mm}
\noindent\textbf{DAVIS.} 
Table~\ref{Tab:DAVIS} reports the experimental results on DAVIS 2016 and DAVIS 2017. Besides existing algorithms, We also evaluate the classical memory-based methods (STM~\cite{STM-VOS} and HMMNet~\cite{HMMNet-VOS}) and the powerful self-supervised method (MAST~\cite{MAST-VOS}) in the scribble/box-supervised manner. We train the STM and HMMNet models using the scribble annotations with our loss function $\mathcal L$, and the resulting models are referred to as SS-STM and SS-HMMNet, respectively. We convert the box and the scribble to the dense mask with the box encoder~\cite{LWL-VOS} and GrabCut~\cite{GrabCut}, respectively, and then use the generated dense mask to initialize MAST. These two variants of MAST are called MAST-Box and MAST-Scribble.

Compared to SS-STM and SS-HMMNet, our model obtains considerable performance gains of 8.7\%/3.5\% and 14.1\%/7.9\% in $\mathcal{J}\&\mathcal{F}$ on DAVIS 2016 and 2017, respectively, which validates the effectiveness of our RHMNet. The large performance gaps between MAST and MAST-Box/MAST-Scribble demonstrate that MAST is sensitive to the quality of the initial mask. It shows that directly reforming a mask-initialized VOS model, such as MAST, to a box/scribble-initialized one is sub-optimal. 
Compared with MAST, which requires precise mask annotation for initialization, our RHMNet achieves a performance gain of 1.9\% in $\mathcal{J}\&\mathcal{F}$ on DAVIS 2017. It shows that our method can achieve a good balance between performance and annotation labor.
Our method also performs comparably with the box-initialized methods, such as E3SN, SRCNN, and LWL-Box, which require massive mask annotations for offline training. It demonstrates that the scribble is an effective weak annotation form for VOS.
Huang \etal~\cite{SSVOS} achieve a score of 55.7\% in $\mathcal{J}\&\mathcal{F}$ on DAVIS 2017 using their own scribble annotations. Although we cannot directly compare their algorithm with ours, the substantial performance gap reflects the effectiveness of our method.

\begin{table}[t]
\setlength\tabcolsep{2.0pt}
\begin{center}
\caption{Experimental results on the Youtube-VOS 2018 validation set.}
\renewcommand\arraystretch{1.1}
\resizebox{1.0\linewidth}{!}{
\begin{tabular}{l|c|c|ccccc}
\toprule
Methods & Init. Anno. & Training Label & $\mathcal{G}$ & $\mathcal{J_{S}}$ & $\mathcal{F_{S}}$ & $\mathcal{J_{U}}$ & $\mathcal{F_{U}}$\\
\midrule
RGMP~\cite{RGMP-VOS}    & Mask & Mask & 53.8 & 59.5 & --   & 45.2 & --   \\
A-GAME~\cite{GAM-VOS}   & Mask & Mask & 66.1 & 67.8 & --   & 60.8 & --   \\
STM~\cite{STM-VOS}      & Mask & Mask & 79.4 & 79.7 & 84.2 & 72.8 & 80.9 \\
CFBI~\cite{CFBI-VOS}    & Mask & Mask & 81.4 & 81.1 & 85.8 & 75.3 & 83.4 \\
LWL~\cite{LWL-VOS}      & Mask & Mask & 81.5 & 80.4 & 84.9 & 76.4 & 84.4 \\
HMMNet~\cite{HMMNet-VOS}& Mask & Mask & \textbf{82.6} & \textbf{82.1} & \textbf{87.0} & \textbf{76.8} & \textbf{86.4} \\
\midrule
SiamMask~\cite{SiamMask}& Box  & Mask & 52.8 & 60.2 & 58.2 & 45.1 & 47.7 \\
QMA~\cite{QMRA-VOS}     & Box  & Mask & 67.6 & 71.2   & --   & 58.1   & --   \\
SRCNN~\cite{SiamRCNN}& Box  & Mask & 68.3 & 69.9 & --   & 61.4 & --   \\
LWL-Box~\cite{LWL-VOS}  & Box  & Mask & \textbf{70.2} & \textbf{72.7} & \textbf{75.1} & \textbf{62.5} & \textbf{70.4} \\
\midrule
Vid.Color~\cite{VidColor}       & Mask & None & 38.9 & 43.1 & 38.6 & 36.6 & 37.4 \\
CorrFlow~\cite{CorrFlow}        & Mask & None & 46.6 & 50.6 & 46.6 & 43.8 & 45.6 \\
MAST~\cite{MAST-VOS}            & Mask & None & \textbf{64.2} & \textbf{63.9} & \textbf{64.9} & \textbf{60.3} & \textbf{67.7} \\
\midrule
MAST-Box    & Box & Mask & 54.4 & 58.0 & 57.9 & 48.5 & 53.3 \\
MAST-Scribble    & Scribble & None & 35.1 & 38.3 & 37.0 & 30.9 & 34.4 \\
SS-STM         & Scribble & Scribble & 44.4 & 49.6 & 49.7 & 37.0 & 41.4\\
SS-HMMNet        & Scribble & Scribble & 43.6 & 49.9 & 49.1 & 36.5 & 39.1\\
Huang \etal~\cite{SSVOS}   & Scribble$^*$ & Scribble$^*$ & 51.1 & 53.1 & 51.8 & 47.3 & 52.0\\
Ours            & Scribble & Scribble & \textbf{59.6} & \textbf{64.9} & \textbf{65.8} & \textbf{51.3} & \textbf{56.6} \\
\bottomrule
\end{tabular}}
\label{Tab:Youtube-VOS2018}
\end{center}
\end{table}

\vspace{1mm}
\noindent\textbf{Youtube-VOS.}
Table~\ref{Tab:Youtube-VOS2018} reports the experimental results on the Youtube-VOS 2018 validation set. Compared with SS-STM and SS-HMMNet, our method obtains performance gains of 15.2\% and 16.0\% in $\mathcal{G}$, respectively, which validates the effectiveness of our RHMNet. Huang \etal~\cite{SSVOS} obtain a score of 51.1\% in $\mathcal{G}$ while our method achieves a substantially higher score of 59.6\% in $\mathcal{G}$. Besides, our approach also outperforms MASK-Scribble and Mask-Box. Compared with MAST, our method performs marginally better on seen categories but obtains inferior performance on unseen categories. As a self-supervised algorithm, MAST achieves excellent generalization ability to unseen objects, while our RHMNet shows limited generalization. Despite the limitation in generalization, our method still performs better than the other scribble-initialized methods on unseen objects. In addition, the superior performance of our method on seen categories demonstrates the potential of our method if providing more data with low-cost scribble annotation for training.

Table~\ref{Tab:Youtube-VOS2019} presents the results on the Youtube-VOS 2019 validation set. Compared with SS-STM and SS-HMMNet, our approach achieves performance gains of 8.1\% and 6.0\% in $\mathcal{G}$, which manifests the effectiveness of our method.

\begin{table}[t]
\setlength\tabcolsep{2.0pt}
\begin{center}
\caption{Experimental results on the Youtube-VOS 2019 validation set.}
\label{Tab:Youtube-VOS2019}
\scriptsize
\renewcommand\arraystretch{1.05}
\resizebox{1.0\linewidth}{!}{
\begin{tabular}{l|c|c|ccccc}
\toprule
Methods & Init. Anno. & Training Label & $\mathcal{G}$ & $\mathcal{J_{S}}$ & $\mathcal{F_{S}}$ & $\mathcal{J_{U}}$ & $\mathcal{F_{U}}$\\
\midrule
STM~\cite{STM-VOS}       & Mask & Mask & 79.3 & 79.8 & 83.8 & 73.0 & 80.5 \\
KMN~\cite{KMN-VOS}       & Mask & Mask & 80.0 & 80.4 & 84.5 & 73.8 & 81.4 \\
LWL~\cite{LWL-VOS}       & Mask & Mask & 81.0 & 79.6 & 83.8 & 76.4 & 84.2 \\
CFBI~\cite{CFBI-VOS}     & Mask & Mask & 81.0 & 80.6 & 85.1 & 75.2 & 83.0 \\
HMMNet~\cite{HMMNet-VOS} & Mask & Mask & \textbf{82.5} & \textbf{81.7} & \textbf{86.1} & \textbf{77.3} & \textbf{85.0} \\
\midrule
MAST-Box    & Box & Mask & 54.2 & 57.2 & 56.7 & 49.3 & 53.8 \\
MAST-Scribble    & Scribble & None & 35.3 & 39.0 & 37.0 & 31.0 & 34.3 \\
SS-STM          & Scribble & Scribble & 41.2 & 46.8 & 47.2 & 33.3 & 37.7 \\
SS-HMMNet          & Scribble & Scribble & 43.3 & 49.3 & 48.2 & 36.7 & 39.2 \\
Ours             & Scribble & Scribble & \textbf{59.3} & \textbf{63.9} & \textbf{64.8} & \textbf{51.7} & \textbf{56.7} \\
\bottomrule
\end{tabular}}
\end{center}
\end{table}

\begin{figure*}[t]
\centering
    \includegraphics[width=1.0\textwidth]{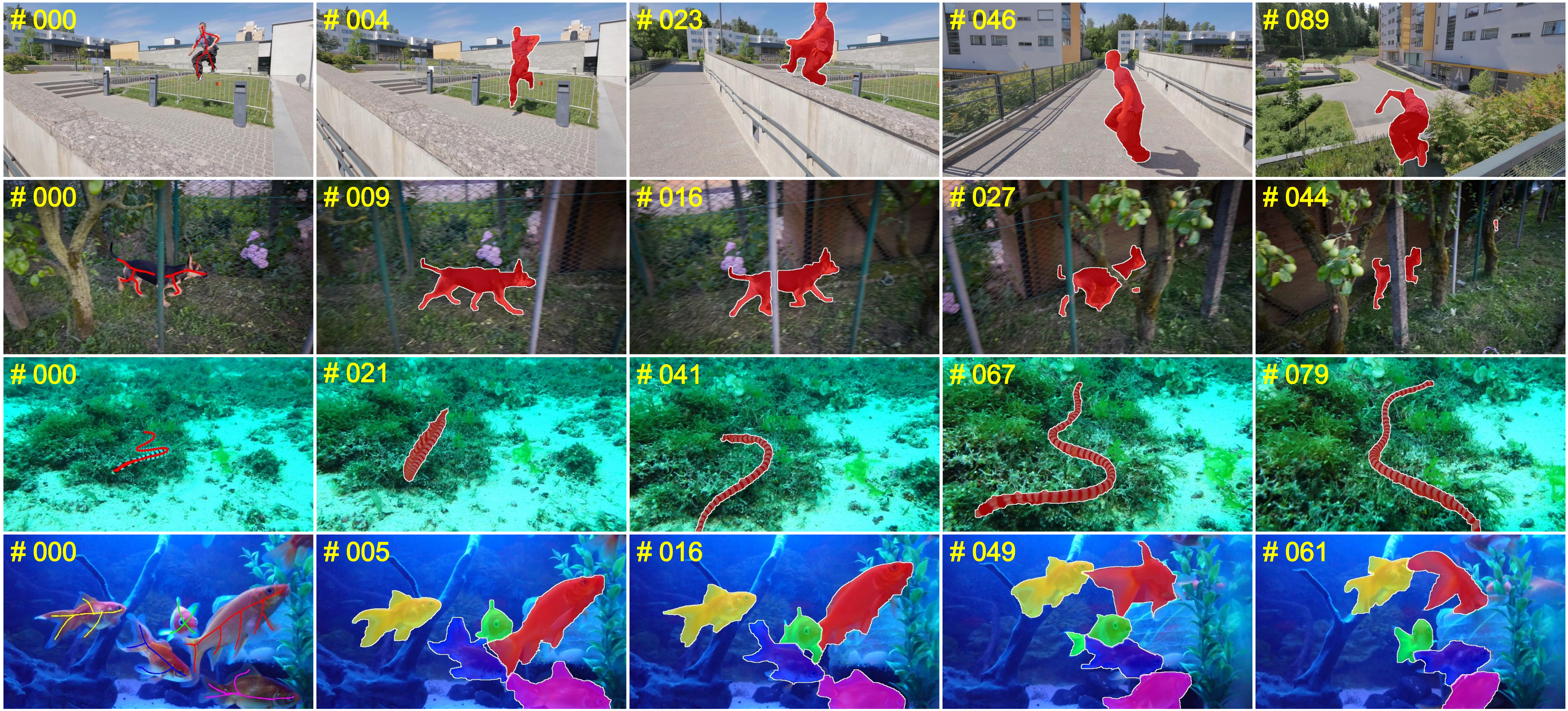}
    \caption{Qualitative results on four challenging sequences. The first column shows the initial target scribbles. Our method produces accurate segmentation in these challenging situations.
    }
\label{Fig:Visualization_mask}
\end{figure*}

\subsection{Qualitative results}
Figure~\ref{Fig:Visualization_mask} shows the qualitative results of our algorithm on four challenging sequences. The initial target scribbles are shown in the first column. We can observe that our method succeeds in dealing with kinds of challenges, including complex motion (the first row), occlusion (the second row), winding object (the third row), and similar targets belonging to the same category (the last row). These favorable results demonstrate the substantial potential of our proposed RHMNet to reduce the annotation burden of users for representing the desired object in human-machine interaction.

\subsection{Case studies}
To further obtain more insights into the pros and cons of our proposed RHMNet, we conduct more case studies on several challenging situations.
\begin{figure}[t]
\centering
    \includegraphics[width=1.0\columnwidth]{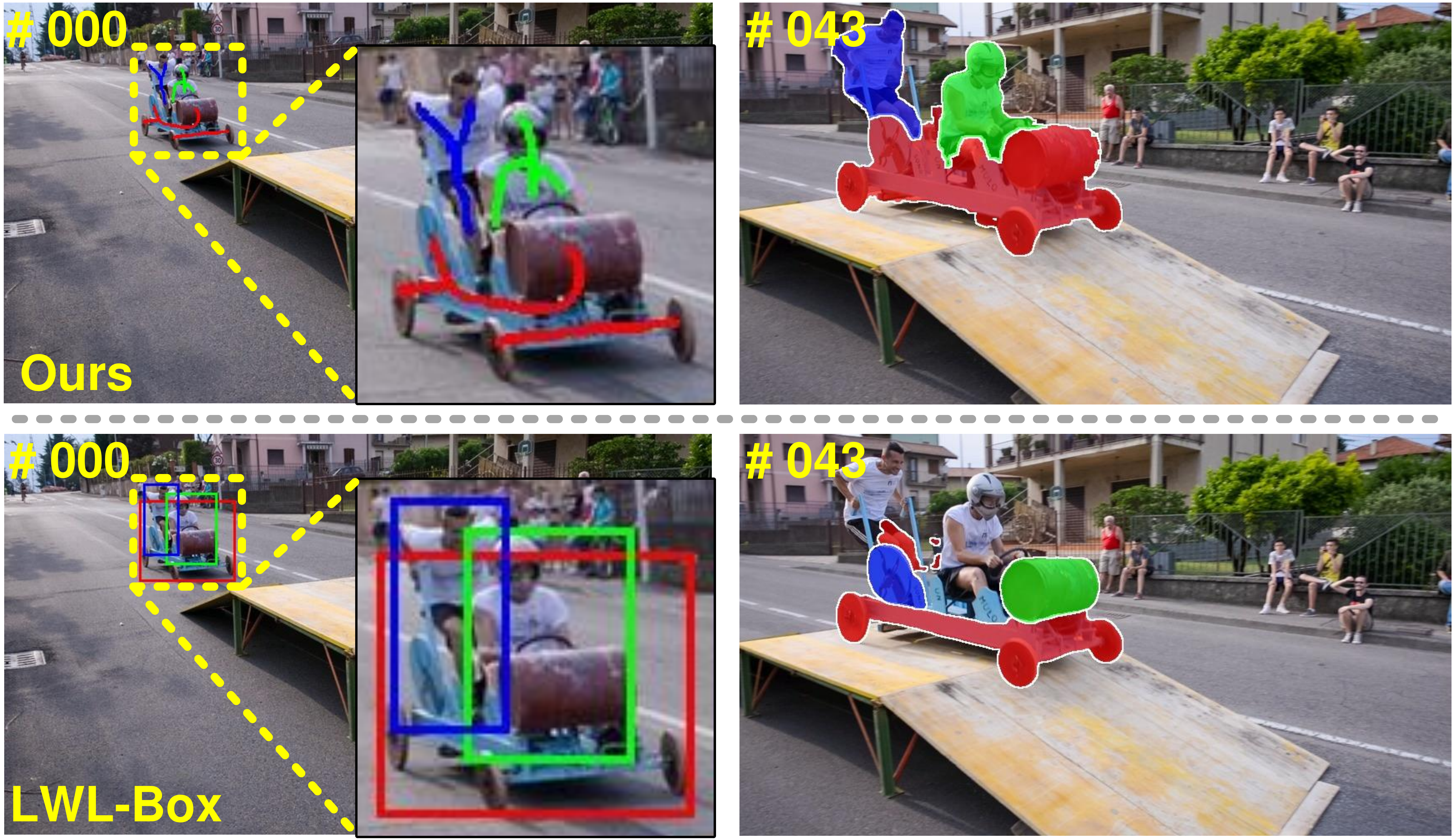}
    \caption{Comparison between the segmentation results of using bounding boxes and scribbles for initialization. The bounding box is ambiguous and damages the segmentation performance when the objects occlude each other partially.}
\label{Fig:Compare_with_bbox}
\end{figure}

\begin{figure}[t]
\centering
    \includegraphics[width=1.0\columnwidth]{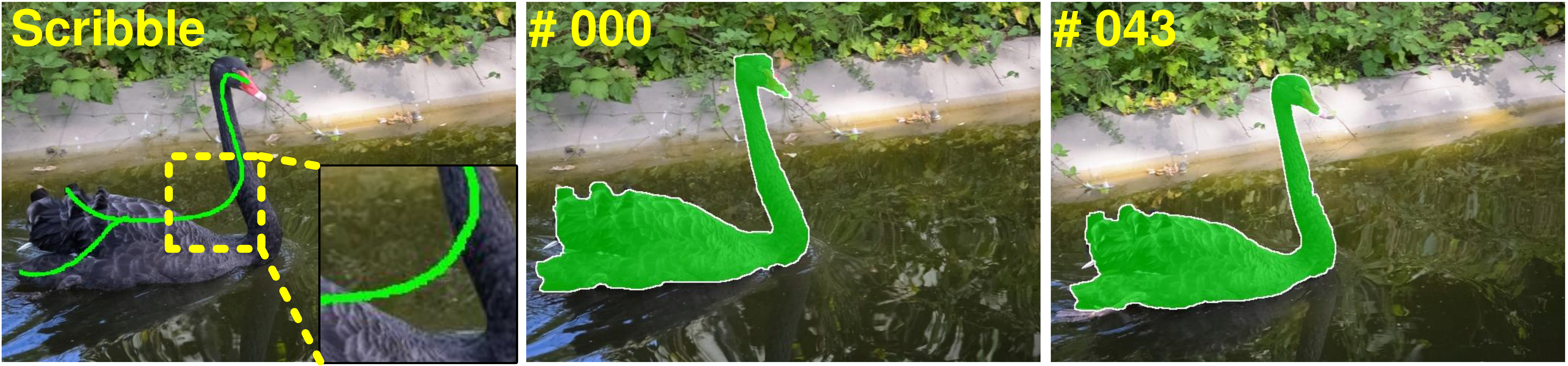}
    \caption{Segmentation result of our algorithm in the case where part of the initial target scribble is out of target.}
\label{Fig:Robust_to_scribble}
\end{figure}

\begin{figure}[t]
\centering
    \includegraphics[width=1.0\columnwidth]{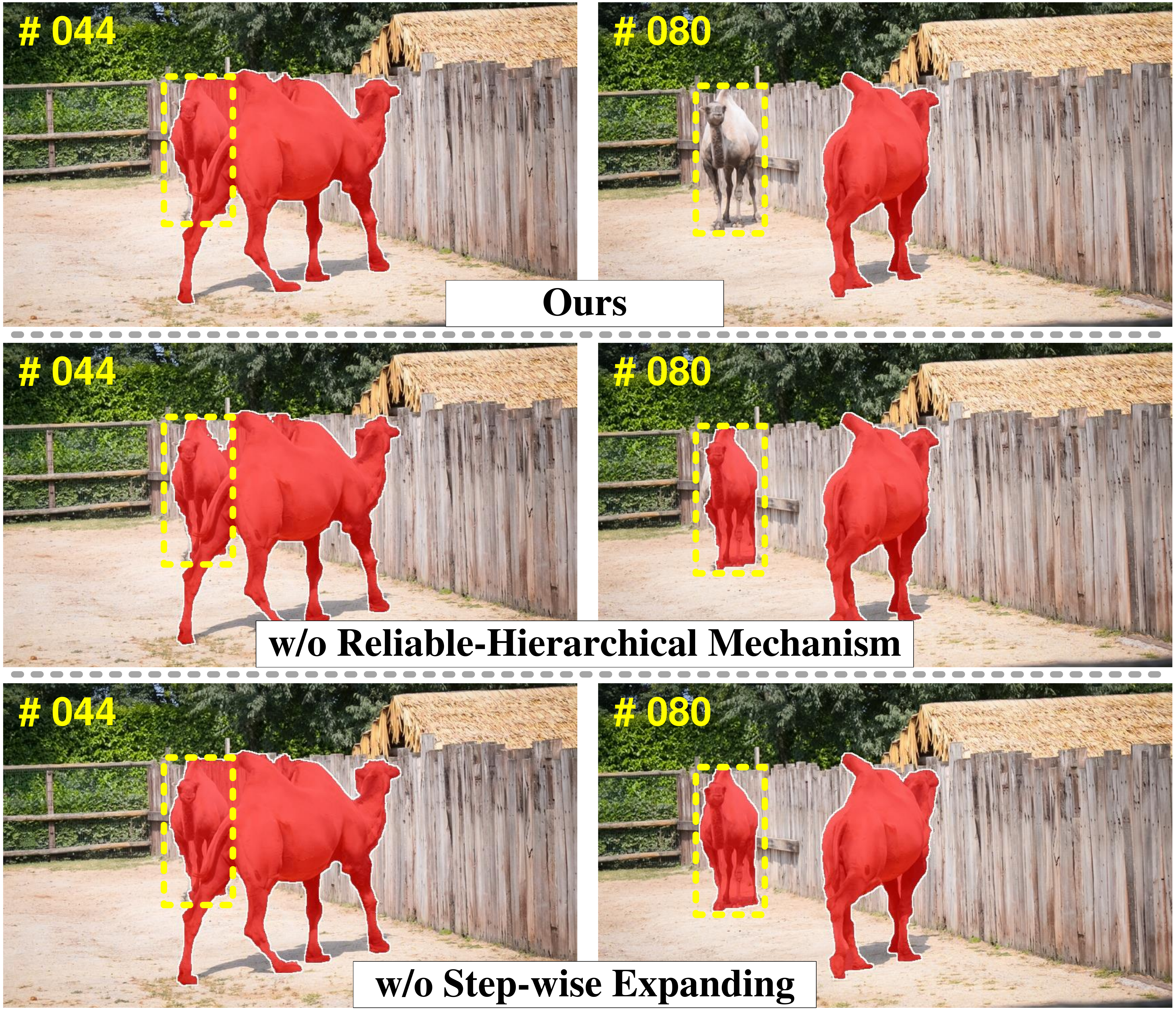}
    \caption{Comparison between the segmentation results of our method and two variants without the reliable-hierarchical mechanism and the step-wise expanding strategy, respectively. Compared with our method, these two variants are much more sensitive to the previous prediction errors.}
\label{Fig:Reliable_region}
\end{figure}

\vspace{1mm}
\noindent\textbf{Comparison with using bounding boxes for initialization.} Compared with the bounding box, which is ambiguous in some cases, the scribble is much more flexible for representing an object. Figure~\ref{Fig:Compare_with_bbox} illustrates the segmentation results of our method and LWL-Box~\cite{LWL-VOS} on a sequence in which the objects occlude each other partially. The partial occlusion results in the bounding box of one object also enclosing part of the other objects, which makes LWL-Box confuse different objects. By contrast, our method understands the scribble annotation correctly and produces more accurate segmentation masks.

\vspace{1mm}
\noindent\textbf{Robustness to inaccurate scribbles.} 
A user-friendly SS-VOS algorithm is supposed to be robust to the scribble containing some unexpected inaccurate parts. Figure~\ref{Fig:Robust_to_scribble} shows the case where part of the initial target scribble is out of the target. We can observe that our method still predicts an accurate target mask. We attribute the robustness to our similarity-based scribble downsampling module, which adaptively assigns a low weight for the inaccurate part of the initial scribble.

\vspace{1mm}
\noindent\textbf{Effect of the reliable-hierarchical mechanism and step-wise expanding strategy.}
To visually analyze the effect of the reliable-hierarchical mechanism and step-wise expanding strategy, we show the segmentation results of the variants without using them in Figure~\ref{Fig:Reliable_region}. In the $44^{th}$ frame, when the target camel overlaps with the background camel (distractor), all the variants and our method inevitably misrecognize the distractor as the target. When the target camel moves far away from the background camel (the $80^{th}$ frame), our method successfully distinguishes the target from the distractor, demonstrating that our approach is robust to previous prediction errors. By contrast, the other two variants still misrecognize the background camel as the target camel. Namely, for the two variants, the previous prediction errors dominate the future prediction. The visual comparisons validate the effectiveness of the reliable-hierarchical mechanism and the step-wise expanding strategy.

\section{Conclusion}
In this paper, we investigate the problem of scribble-supervised video object segmentation, which can sufficiently lighten the annotation burden for both offline training and online initialization. 
A reliability-hierarchical memory network is proposed to perform video object segmentation only knowing the target information denoted by a sparse scribble. It works in a step-wise expanding strategy w.r.t. the memory reliability level to robustly predict the dense mask of the target, effectively mitigating the effect of the previous prediction errors. In addition, a scribble-supervised learning mechanism mining the pixel-level relation and frame-level relation is proposed to effectively train the model to predict dense segmentation results using only sparse scribble annotations. The favorable performance on two popular benchmarks demonstrates that the proposed approach is promising.

\section{Acknowledgements}
This study was supported in part by the Key Research Project of Peng Cheng Laboratory (PCL2021A07), in part by the National Natural Science Foundation of China (62172126), and in part by the Shenzhen Research Council (JCYJ20210324120202006 and JCYJ20210324132212030).

\bibliographystyle{IEEEtran}
\bibliography{ssvos}

\begin{thebibliography}{10}
\providecommand{\url}[1]{#1}
\csname url@samestyle\endcsname
\providecommand{\newblock}{\relax}
\providecommand{\bibinfo}[2]{#2}
\providecommand{\BIBentrySTDinterwordspacing}{\spaceskip=0pt\relax}
\providecommand{\BIBentryALTinterwordstretchfactor}{4}
\providecommand{\BIBentryALTinterwordspacing}{\spaceskip=\fontdimen2\font plus
\BIBentryALTinterwordstretchfactor\fontdimen3\font minus
  \fontdimen4\font\relax}
\providecommand{\BIBforeignlanguage}[2]{{%
\expandafter\ifx\csname l@#1\endcsname\relax
\typeout{** WARNING: IEEEtran.bst: No hyphenation pattern has been}%
\typeout{** loaded for the language `#1'. Using the pattern for}%
\typeout{** the default language instead.}%
\else
\language=\csname l@#1\endcsname
\fi
#2}}
\providecommand{\BIBdecl}{\relax}
\BIBdecl

\bibitem{Robot-Manipulation}
A.~Abramov, K.~Pauwels, J.~Papon, F.~W{\"o}rg{\"o}tter, and B.~Dellen,
  ``Depth-supported real-time video segmentation with the kinect,'' in
  \emph{Proceedings of the IEEE Workshop on the Applications of Computer Vision
  (WACV)}, 2012.

\bibitem{Robot-Grasps}
H.~Kjellstrom, J.~Romero, and D.~Kragic, ``Visual recognition of grasps for
  human-to-robot mapping,'' in \emph{Proceedings of the IEEE/RSJ International
  Conference on Intelligent Robots and Systems}, 2008.

\bibitem{Surveillance}
I.~Cohen and G.~Medioni, ``Detecting and tracking moving objects for video
  surveillance,'' in \emph{Proceedings of the IEEE Computer Society Conference
  on Computer Vision and Pattern Recognition (Cat. No PR00149)}, 1999.

\bibitem{Camera}
A.~Erd{\'e}lyi, T.~Bar{\'a}t, P.~Valet, T.~Winkler, and B.~Rinner, ``Adaptive
  cartooning for privacy protection in camera networks,'' in \emph{Proceedings
  of the IEEE International Conference on Advanced Video and Signal Based
  Surveillance (AVSS)}, 2014.

\bibitem{URVOS-TIP}
P.~W. Patil, A.~Dudhane, A.~Kulkarni, S.~Murala, A.~B. Gonde, and S.~Gupta,
  ``An unified recurrent video object segmentation framework for various
  surveillance environments,'' \emph{IEEE Transactions on Image Processing},
  vol.~30, pp. 7889--7902, 2021.

\bibitem{VOS-RMP}
S.~W. Oh, J.-Y. Lee, K.~Sunkavalli, and S.~J. Kim, ``Fast video object
  segmentation by reference-guided mask propagation,'' in \emph{Proceedings of
  the IEEE/CVF Conference on Computer Vision and Pattern Recognition}, 2018.

\bibitem{LITL-VOS}
Y.~Mao, N.~Wang, W.~Zhou, and H.~Li, ``Joint inductive and transductive
  learning for video object segmentation,'' in \emph{Proceedings of the
  IEEE/CVF International Conference on Computer Vision}, 2021.

\bibitem{STM-VOS}
S.~W. Oh, J.-Y. Lee, N.~Xu, and S.~J. Kim, ``Video object segmentation using
  space-time memory networks,'' in \emph{Proceedings of the IEEE/CVF
  International Conference on Computer Vision}, 2019.

\bibitem{PClip-VOS}
K.~Park, S.~Woo, S.~W. Oh, I.~S. Kweon, and J.-Y. Lee, ``Per-clip video object
  segmentation,'' in \emph{Proceedings of the IEEE/CVF Conference on Computer
  Vision and Pattern Recognition}, 2022.

\bibitem{XMem-VOS}
H.~K. Cheng and A.~G. Schwing, ``Xmem: Long-term video object segmentation with
  an atkinson-shiffrin memory model,'' in \emph{Proceedings of the European
  Conference on Computer Vision}, 2022.

\bibitem{Unicorn}
B.~Yan, Y.~Jiang, P.~Sun, D.~Wang, Z.~Yuan, P.~Luo, and H.~Lu, ``Towards grand
  unification of object tracking,'' in \emph{Proceedings of the European
  Conference on Computer Vision}.\hskip 1em plus 0.5em minus 0.4em\relax
  Springer, 2022, pp. 733--751.

\bibitem{LWL-VOS}
G.~Bhat, F.~J. Lawin, M.~Danelljan, A.~Robinson, M.~Felsberg, L.~V. Gool, and
  R.~Timofte, ``Learning what to learn for video object segmentation,'' in
  \emph{Proceedings of the European Conference on Computer Vision}, 2020.

\bibitem{KMN-VOS}
H.~Seong, J.~Hyun, and E.~Kim, ``Kernelized memory network for video object
  segmentation,'' in \emph{Proceedings of the European Conference on Computer
  Vision}, 2020.

\bibitem{RANet}
Z.~Wang, J.~Xu, L.~Liu, F.~Zhu, and L.~Shao, ``Ranet: Ranking attention network
  for fast video object segmentation,'' in \emph{Proceedings of the IEEE/CVF
  International Conference on Computer Vision}, 2019.

\bibitem{AOTrans}
Z.~Yang, Y.~Wei, and Y.~Yang, ``Associating objects with transformers for video
  object segmentation,'' in \emph{Proceedings of the Conference on Advances in
  Neural Information Processing Systems}, 2021.

\bibitem{WSDDet}
H.~Bilen and A.~Vedaldi, ``Weakly supervised deep detection networks,'' in
  \emph{Proceedings of the IEEE/CVF Conference on Computer Vision and Pattern
  Recognition}, 2016.

\bibitem{ResWSOD}
Y.~Shen, R.~Ji, Y.~Wang, Z.~Chen, F.~Zheng, F.~Huang, and Y.~Wu, ``Enabling
  deep residual networks for weakly supervised object detection,'' in
  \emph{Proceedings of the European Conference on Computer Vision}, 2020.

\bibitem{Scribblesup}
D.~Lin, J.~Dai, J.~Jia, K.~He, and J.~Sun, ``Scribblesup: Scribble-supervised
  convolutional networks for semantic segmentation,'' in \emph{Proceedings of
  the IEEE/CVF Conference on Computer Vision and Pattern Recognition}, 2016.

\bibitem{CPWSSS}
F.~Zhang, C.~Gu, C.~Zhang, and Y.~Dai, ``Complementary patch for weakly
  supervised semantic segmentation,'' in \emph{Proceedings of the IEEE/CVF
  International Conference on Computer Vision}, 2021.

\bibitem{SSVOS}
P.~Huang, J.~Han, N.~Liu, J.~Ren, and D.~Zhang, ``Scribble-supervised video
  object segmentation,'' \emph{IEEE/CAA Journal of Automatica Sinica}, vol.~9,
  no.~2, pp. 339--353, 2021.

\bibitem{GAM-VOS}
J.~Johnander, M.~Danelljan, E.~Brissman, F.~S. Khan, and M.~Felsberg, ``A
  generative appearance model for end-to-end video object segmentation,'' in
  \emph{Proceedings of the IEEE/CVF Conference on Computer Vision and Pattern
  Recognition}, 2019.

\bibitem{DAVIS}
S.~Caelles, J.~Pont-Tuset, F.~Perazzi, A.~Montes, K.-K. Maninis, and L.~{Van
  Gool}, ``The 2019 davis challenge on vos: Unsupervised multi-object
  segmentation,'' \emph{arXiv:1905.00737}, 2019.

\bibitem{Youtube-vos}
N.~Xu, L.~Yang, Y.~Fan, J.~Yang, D.~Yue, Y.~Liang, B.~Price, S.~Cohen, and
  T.~Huang, ``Youtube-vos: Sequence-to-sequence video object segmentation,'' in
  \emph{Proceedings of the European Conference on Computer Vision}, 2018.

\bibitem{SAT-VOS}
X.~Chen, Z.~Li, Y.~Yuan, G.~Yu, J.~Shen, and D.~Qi, ``State-aware tracker for
  real-time video object segmentation,'' in \emph{Proceedings of the IEEE/CVF
  Conference on Computer Vision and Pattern Recognition}, 2020.

\bibitem{MaskRCNN}
Y.-T. Hu, J.-B. Huang, and A.~Schwing, ``Maskrnn: Instance level video object
  segmentation,'' in \emph{Proceedings of the Conference on Advances in Neural
  Information Processing Systems}, 2017.

\bibitem{RAMP-VOS}
X.~Li and C.~C. Loy, ``Video object segmentation with joint re-identification
  and attention-aware mask propagation,'' in \emph{Proceedings of the European
  Conference on Computer Vision}, 2018.

\bibitem{LFSI-VOS}
F.~Perazzi, A.~Khoreva, R.~Benenson, B.~Schiele, and A.~Sorkine-Hornung,
  ``Learning video object segmentation from static images,'' in
  \emph{Proceedings of the IEEE/CVF Conference on Computer Vision and Pattern
  Recognition}, 2017.

\bibitem{TA-VOS}
Y.~Zhang, Z.~Wu, H.~Peng, and S.~Lin, ``A transductive approach for video
  object segmentation,'' in \emph{Proceedings of the IEEE/CVF Conference on
  Computer Vision and Pattern Recognition}, 2020.

\bibitem{VM-VOS}
Y.-T. Hu, J.-B. Huang, and A.~G. Schwing, ``Videomatch: Matching based video
  object segmentation,'' in \emph{Proceedings of the European Conference on
  Computer Vision}, 2018.

\bibitem{DTM-VOS}
X.~Huang, J.~Xu, Y.-W. Tai, and C.-K. Tang, ``Fast video object segmentation
  with temporal aggregation network and dynamic template matching,'' in
  \emph{Proceedings of the IEEE/CVF Conference on Computer Vision and Pattern
  Recognition}, 2020.

\bibitem{PLM-VOS}
J.~Shin~Yoon, F.~Rameau, J.~Kim, S.~Lee, S.~Shin, and I.~So~Kweon,
  ``Pixel-level matching for video object segmentation using convolutional
  neural networks,'' in \emph{Proceedings of the IEEE/CVF International
  Conference on Computer Vision}, 2017.

\bibitem{CFBI-VOS}
Z.~Yang, Y.~Wei, and Y.~Yang, ``Collaborative video object segmentation by
  foreground-background integration,'' in \emph{Proceedings of the European
  Conference on Computer Vision}, 2020.

\bibitem{ASRF-TIP}
L.~Hong, W.~Zhang, L.~Chen, W.~Zhang, and J.~Fan, ``Adaptive selection of
  reference frames for video object segmentation,'' \emph{IEEE Transactions on
  Image Processing}, vol.~31, pp. 1057--1071, 2021.

\bibitem{DMM-VOS}
X.~Zeng, R.~Liao, L.~Gu, Y.~Xiong, S.~Fidler, and R.~Urtasun, ``Dmm-net:
  Differentiable mask-matching network for video object segmentation,'' in
  \emph{Proceedings of the IEEE/CVF International Conference on Computer
  Vision}, 2019.

\bibitem{TacklingBD}
S.~Cho, H.~Lee, M.~Lee, C.~Park, S.~Jang, M.~Kim, and S.~Lee, ``Tackling
  background distraction in video object segmentation,'' in \emph{Proceedings
  of the European Conference on Computer Vision}.\hskip 1em plus 0.5em minus
  0.4em\relax Springer, 2022, pp. 446--462.

\bibitem{EGMN-VOS}
X.~Lu, W.~Wang, M.~Danelljan, T.~Zhou, J.~Shen, and L.~V. Gool, ``Video object
  segmentation with episodic graph memory networks,'' in \emph{Proceedings of
  the European Conference on Computer Vision}, 2020.

\bibitem{MaskVOS-TIP}
M.~Wang, J.~Mei, L.~Liu, G.~Tian, Y.~Liu, and Z.~Pan, ``Delving deeper into
  mask utilization in video object segmentation,'' \emph{IEEE Transactions on
  Image Processing}, vol.~31, pp. 6255--6266, 2022.

\bibitem{AOLM-TIP}
P.~Guo, W.~Zhang, X.~Li, and W.~Zhang, ``Adaptive online mutual learning
  bi-decoders for video object segmentation,'' \emph{IEEE Transactions on Image
  Processing}, vol.~31, pp. 7063--7077, 2022.

\bibitem{HMMNet-VOS}
H.~Seong, S.~W. Oh, J.-Y. Lee, S.~Lee, S.~Lee, and E.~Kim, ``Hierarchical
  memory matching network for video object segmentation,'' in \emph{Proceedings
  of the IEEE/CVF International Conference on Computer Vision}, 2021.

\bibitem{QADM-VOS}
Y.~Liu, R.~Yu, F.~Yin, X.~Zhao, W.~Zhao, W.~Xia, and Y.~Yang, ``Learning
  quality-aware dynamic memory for video object segmentation,'' in
  \emph{Proceedings of the European Conference on Computer Vision}, 2022.

\bibitem{SWEM}
Z.~Lin, T.~Yang, M.~Li, Z.~Wang, C.~Yuan, W.~Jiang, and W.~Liu, ``Swem: Towards
  real-time video object segmentation with sequential weighted
  expectation-maximization,'' in \emph{Proceedings of the IEEE/CVF Conference
  on Computer Vision and Pattern Recognition}, 2022, pp. 1362--1372.

\bibitem{QMRA-VOS}
F.~Lin, H.~Xie, Y.~Li, and Y.~Zhang, ``Query-memory re-aggregation for
  weakly-supervised video object segmentation,'' in \emph{Proceedings of the
  AAAI Conference on Artificial Intelligence}, 2021.

\bibitem{SiamRCNN}
P.~Voigtlaender, J.~Luiten, P.~H. Torr, and B.~Leibe, ``Siam r-cnn: Visual
  tracking by re-detection,'' in \emph{Proceedings of the IEEE/CVF Conference
  on Computer Vision and Pattern Recognition}, 2020.

\bibitem{SiamMask}
Q.~Wang, L.~Zhang, L.~Bertinetto, W.~Hu, and P.~H. Torr, ``Fast online object
  tracking and segmentation: A unifying approach,'' in \emph{Proceedings of the
  IEEE/CVF Conference on Computer Vision and Pattern Recognition}, 2019.

\bibitem{E3SN}
M.~Lan, Y.~Zhang, Q.~Xu, and L.~Zhang, ``E3sn: efficient end-to-end siamese
  network for video object segmentation,'' in \emph{Proceedings of the
  International Joint Conference on Artificial Intelligence}, 2021.

\bibitem{MIVOS}
H.~K. Cheng, Y.-W. Tai, and C.-K. Tang, ``Modular interactive video object
  segmentation: Interaction-to-mask, propagation and difference-aware fusion,''
  in \emph{Proceedings of the IEEE/CVF Conference on Computer Vision and
  Pattern Recognition}, 2021.

\bibitem{UGVOS}
S.~W. Oh, J.-Y. Lee, N.~Xu, and S.~J. Kim, ``Fast user-guided video object
  segmentation by interaction-and-propagation networks,'' in \emph{Proceedings
  of the IEEE/CVF Conference on Computer Vision and Pattern Recognition}, 2019.

\bibitem{GatedCRF-WSSS}
A.~Obukhov, S.~Georgoulis, D.~Dai, and L.~Van~Gool, ``Gated crf loss for weakly
  supervised semantic image segmentation,'' \emph{arXiv preprint
  arXiv:1906.04651}, 2019.

\bibitem{NormalizedCut-WSSS}
M.~Tang, A.~Djelouah, F.~Perazzi, Y.~Boykov, and C.~Schroers, ``Normalized cut
  loss for weakly-supervised cnn segmentation,'' in \emph{Proceedings of the
  IEEE/CVF Conference on Computer Vision and Pattern Recognition}, 2018.

\bibitem{KernelCut-WSSS}
M.~Tang, F.~Perazzi, A.~Djelouah, I.~Ben~Ayed, C.~Schroers, and Y.~Boykov, ``On
  regularized losses for weakly-supervised cnn segmentation,'' in
  \emph{Proceedings of the European Conference on Computer Vision}, 2018.

\bibitem{BPG-WSSS}
B.~Wang, G.~Qi, S.~Tang, T.~Zhang, Y.~Wei, L.~Li, and Y.~Zhang, ``Boundary
  perception guidance: A scribble-supervised semantic segmentation approach,''
  in \emph{Proceedings of the International Joint Conference on Artificial
  Intelligence}, 2019.

\bibitem{DFR-WSSS}
B.~Zhang, J.~Xiao, and Y.~Zhao, ``Dynamic feature regularized loss for weakly
  supervised semantic segmentation,'' \emph{arXiv preprint arXiv:2108.01296},
  2021.

\bibitem{SCWSSOD}
S.~Yu, B.~Zhang, J.~Xiao, and E.~G. Lim, ``Structure-consistent weakly
  supervised salient object detection with local saliency coherence,'' in
  \emph{Proceedings of the AAAI Conference on Artificial Intelligence}, 2021.

\bibitem{WSSA}
J.~Zhang, X.~Yu, A.~Li, P.~Song, B.~Liu, and Y.~Dai, ``Weakly-supervised
  salient object detection via scribble annotations,'' in \emph{Proceedings of
  the IEEE/CVF Conference on Computer Vision and Pattern Recognition}, 2020.

\bibitem{WSVSOD}
W.~Zhao, J.~Zhang, L.~Li, N.~Barnes, N.~Liu, and J.~Han, ``Weakly supervised
  video salient object detection,'' in \emph{Proceedings of the IEEE/CVF
  Conference on Computer Vision and Pattern Recognition}, 2021.

\bibitem{Transformer}
A.~Vaswani, N.~Shazeer, N.~Parmar, J.~Uszkoreit, L.~Jones, A.~N. Gomez,
  {\L}.~Kaiser, and I.~Polosukhin, ``Attention is all you need,'' in
  \emph{Proceedings of the Conference on Advances in Neural Information
  Processing Systems}, 2017.

\bibitem{FocalLoss}
T.-Y. Lin, P.~Goyal, R.~Girshick, K.~He, and P.~Doll{\'a}r, ``Focal loss for
  dense object detection,'' in \emph{Proceedings of the IEEE/CVF International
  Conference on Computer Vision}, 2017.

\bibitem{UMDE}
C.~Godard, O.~Mac~Aodha, and G.~J. Brostow, ``Unsupervised monocular depth
  estimation with left-right consistency,'' in \emph{Proceedings of the
  IEEE/CVF Conference on Computer Vision and Pattern Recognition}, 2017.

\bibitem{ResNet}
K.~He, X.~Zhang, S.~Ren, and J.~Sun, ``Deep residual learning for image
  recognition,'' in \emph{Proceedings of the IEEE/CVF Conference on Computer
  Vision and Pattern Recognition}, 2016.

\bibitem{Imagenet}
J.~Deng, W.~Dong, R.~Socher, L.-J. Li, K.~Li, and L.~Fei-Fei, ``Imagenet: A
  large-scale hierarchical image database,'' in \emph{Proceedings of the
  IEEE/CVF Conference on Computer Vision and Pattern Recognition}, 2009.

\bibitem{COCO}
T.-Y. Lin, M.~Maire, S.~Belongie, J.~Hays, P.~Perona, D.~Ramanan,
  P.~Doll{\'a}r, and C.~L. Zitnick, ``Microsoft coco: Common objects in
  context,'' in \emph{Proceedings of the European Conference on Computer
  Vision}, 2014.

\bibitem{UFO2}
Z.~Ren, Z.~Yu, X.~Yang, M.-Y. Liu, A.~G. Schwing, and J.~Kautz, ``Ufo$^{2}$: A
  unified framework towards omni-supervised object detection,'' in
  \emph{Proceedings of the European Conference on Computer Vision}, 2020.

\bibitem{AdamW}
I.~Loshchilov and F.~Hutter, ``Decoupled weight decay regularization,'' in
  \emph{Proceedings of the International Conference on Learning
  Representations}, 2019.

\bibitem{FEEL-VOS}
P.~Voigtlaender, Y.~Chai, F.~Schroff, H.~Adam, B.~Leibe, and L.-C. Chen,
  ``Feelvos: Fast end-to-end embedding learning for video object
  segmentation,'' in \emph{Proceedings of the IEEE/CVF Conference on Computer
  Vision and Pattern Recognition}, 2019.

\bibitem{RGMP-VOS}
S.~W. Oh, J.-Y. Lee, K.~Sunkavalli, and S.~J. Kim, ``Fast video object
  segmentation by reference-guided mask propagation,'' in \emph{Proceedings of
  the IEEE/CVF Conference on Computer Vision and Pattern Recognition}, 2018.

\bibitem{VidColor}
C.~Vondrick, A.~Shrivastava, A.~Fathi, S.~Guadarrama, and K.~Murphy, ``Tracking
  emerges by colorizing videos,'' in \emph{Proceedings of the European
  Conference on Computer Vision}, 2018.

\bibitem{CycleTime}
X.~Wang, A.~Jabri, and A.~A. Efros, ``Learning correspondence from the
  cycle-consistency of time,'' in \emph{Proceedings of the IEEE/CVF Conference
  on Computer Vision and Pattern Recognition}, 2019.

\bibitem{uvc}
X.~Li, S.~Liu, S.~De~Mello, X.~Wang, J.~Kautz, and M.-H. Yang, ``Joint-task
  self-supervised learning for temporal correspondence,'' in \emph{Proceedings
  of the Conference on tu yAdvances in Neural Information Processing Systems},
  2019.

\bibitem{MAST-VOS}
Z.~Lai, E.~Lu, and W.~Xie, ``Mast: A memory-augmented self-supervised
  tracker,'' in \emph{Proceedings of the IEEE/CVF Conference on Computer Vision
  and Pattern Recognition}, 2020.

\bibitem{GrabCut}
C.~Rother, V.~Kolmogorov, and A.~Blake, ``Grabcut: interactive foreground
  extraction using iterated graph cuts,'' \emph{ACM transactions on graphics
  (TOG)}, 2004.

\bibitem{CorrFlow}
Z.~Lai and W.~Xie, ``Self-supervised learning for video correspondence flow,''
  in \emph{Proceedings of the British Machine Vision Conference}, 2019.

\end{thebibliography}

\newpage

\vfill

\end{document}